\PassOptionsToPackage{dvipsnames,table}{xcolor}
\documentclass[11pt]{article}
\usepackage{acl}

\usepackage[T1]{fontenc}
\usepackage{times}
\usepackage{latexsym}
\usepackage{microtype}
\usepackage{inconsolata}
\usepackage{tikz}
\usetikzlibrary{shapes.geometric, arrows.meta, positioning, fit, backgrounds, calc, shadows}
\usepackage{url}
\usepackage{amsmath,amssymb,amsfonts}
\usepackage{extarrows}
\usepackage{multirow}
\usepackage{enumitem}
\usepackage{tabularx}

\usepackage{multicol}

\usepackage[noend]{algpseudocode}
\usepackage{algorithm}
\usepackage{algorithmicx}
\usepackage{xspace}
\usepackage{dsfont}
\usepackage{epsfig}
\usepackage{epstopdf}
\usepackage{bm}
\usepackage{color}
\usepackage{colortbl}
\usepackage{booktabs} 
\usepackage{tablefootnote}
\usepackage{threeparttable} 
\usepackage{makecell} 
\usepackage{graphicx}
\usepackage{subfig} 

\newcommand{\paratitle}[1]{\vspace{1.1ex}\noindent\textbf{#1}}

\newcommand{\model}{\texttt{SemOPT}\xspace}

\usepackage{tcolorbox}
\tcbuselibrary{skins} 

\newtcolorbox{promptbox}[1][]{
  enhanced,
  width=\textwidth,
  colback=gray!4,
  colframe=gray!45,
  colbacktitle=gray!15,
  coltitle=black,
  boxrule=0.45pt,
  arc=1mm,
  left=6pt, right=6pt, top=6pt, bottom=6pt,
  fontupper=\footnotesize\ttfamily,
  fonttitle=\bfseries\small,
  attach boxed title to top left={xshift=6pt,yshift=-2mm},
  boxed title style={colback=white, colframe=gray!45, boxrule=0.45pt, arc=1mm},
  #1
}

\begin{document}

\title{\model: Fixing Semantic Errors in LLM-based Optimization Modeling via Reward-Guided Search}

\author{
  Zetong Zhou$^{1,2*}$, Wentao Zhang$^{1,2*}$, Jingyuan Wang$^{1,2\dagger}$, Yifan Yang$^{1,2}$, \\
  \textbf{Zizhuo Wang}$^{3}$, \textbf{Shixi Hu}$^{4}$ \\[2pt]
  $^{1}$School of Computer Science and Engineering, Beihang University, Beijing, China \\
  $^{2}$MIIT Key Laboratory of Data and Decision Intelligence, Beihang University, Beijing, China \\
  $^{3}$School of Data Science, The Chinese University of Hong Kong, Shenzhen, China \\
  $^{4}$Cardinal Operations Technology Co., Shanghai, China \\[2pt]
  \texttt{\symbol{123}ztzhou, jywang\symbol{125}@buaa.edu.cn}
}

\maketitle

\renewcommand{\thefootnote}{\fnsymbol{footnote}}
\footnotetext[1]{Equal contribution.}
\footnotetext[2]{Corresponding author.}
\renewcommand{\thefootnote}{\arabic{footnote}}

\begin{abstract}
Operations research supports decision-making in domains such as energy, economics, and healthcare. Solving operations research problems typically begins with optimization modeling, which translates a natural-language problem description into executable solver code. LLMs offer a promising way to automate this process, but they remain prone to errors. In practice, these errors can be divided into two categories: syntactic errors refer to solver code that fails to run successfully or is judged infeasible by the solver; semantic errors refer to solver code that successfully returns an objective value but violates the intent of the original problem. Since semantic errors do not trigger runtime failures, they are difficult to detect and rectify. To address this problem, we introduce \model, a semantic-guided framework for correcting LLM-based optimization models. \model combines a semantic reward model that distinguishes faithful math models from plausible but incorrect ones with an adaptive correction system that applies hierarchical reward-guided search over the modeling space. Experiments on seven optimization modeling benchmarks show that \model establishes a new state of the art and achieves an average 7.6\% accuracy improvement over the strongest baseline on complex datasets.
\end{abstract}

\section{Introduction}

Operations research (OR) provides a fundamental methodology for decision-making in domains such as energy~\cite{energy}, economics~\cite{economics}, healthcare~\cite{math}, and beyond~\cite{others}. Solving OR problems typically starts with optimization modeling, which translates a natural-language problem description into a math model and solver code (e.g., Gurobi)~\cite{huangLLMsMathematicalModeling2025}. Constructing the model traditionally requires substantial domain expertise, but recent progress in LLMs has automated this process, with existing methods using fine-tuning or agent-based workflows to emulate human modeling processes~\cite{jiangLLMOPTLearningDefine2024,huangORLMCustomizableFramework2025,ahmaditeshniziOptiMUSScalableOptimization2024,ahmaditeshniziOptiMUS03UsingLarge2025}.

\begin{figure}[t]
  \centering
  \includegraphics[width=\linewidth]{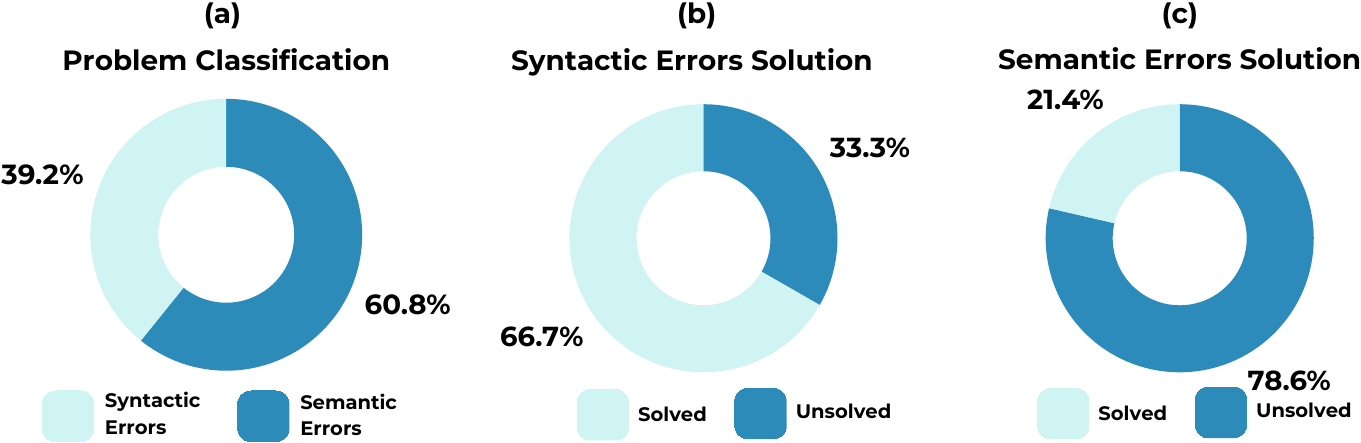}
  \caption{(a) Distribution of syntactic versus semantic errors. (b)(c) Correction rates of syntactic and semantic errors via self-reflection.}
  \label{fig:motivation}
\end{figure}

Despite their promise, LLM-based modeling systems remain error-prone~\cite{yangAutomatedOptimizationModeling2025}. We divide their failures into syntactic and semantic errors. \textit{Syntactic errors} refer to solver code that fails to run successfully or is judged infeasible by the solver. These failures are directly exposed by execution or solver feedback. \textit{Semantic errors}, in contrast, refer to solver code that runs successfully and returns an objective value but violates the intent of the original problem, typically through misunderstood constraints, incorrect objectives, or invalid modeling assumptions. These errors are harder to identify because execution feedback provides no explicit signal of the underlying logical mismatch. To quantify this issue, we prompt an LLM to generate solver code end-to-end on the optimization modeling benchmark IndustryOR~\cite{huangORLMCustomizableFramework2025} and execute the generated codes to categorize failures. As shown in Figure~\ref{fig:motivation}, semantic errors are more prevalent than syntactic errors. Moreover, three rounds of self-reflection~\cite{reflection} correct 66.7\% of syntactic errors but only 21.4\% of semantic errors, confirming that semantic errors are the main bottleneck.

Recent work has started to recognize the importance of correcting semantic errors in optimization modeling, with search-based methods providing the most instructive direction. Autoformulator~\cite{astorgaAutoformulationMathematicalOptimization2025} uses Monte Carlo Tree Search(MCTS) to explore candidate math model components, while SolverLLM~\cite{liSolverLLMLeveragingTestTime2025} incrementally constructs math models over structured elements with outcome-guided search. These methods broaden the hypothesis space beyond single-pass generation, but two limitations remain. First, they often rely on general-purpose LLM evaluators, which are not specifically trained to distinguish semantically faithful math models from plausible but incorrect ones. Second, their search spaces are usually organized around fine-grained mathematical components, entangling high-level modeling decisions with low-level implementation details. As a result, when an error stems from an incorrect modeling strategy, these methods tend to revise local equations rather than backtrack to the reasoning step where the error actually originated.

To overcome these challenges, we propose \model, a semantic-guided framework for correcting LLM-based optimization models. \model first trains a semantic reward model to evaluate the logical faithfulness between a problem description and generated solver code. It then uses this reward signal to guide hierarchical search over three stages: \textit{modeling strategy}, \textit{math model}, and \textit{solver code}, allowing the framework to revise errors at the appropriate level of abstraction. Finally, an adaptive gating mechanism triggers search only when the initial solver code falls below the semantic confidence threshold, reducing unnecessary computation on straightforward instances. Experiments across seven benchmarks show that \model establishes a new state of the art on most datasets.

Our contributions are summarized as follows:
\begin{itemize}[topsep=2pt, itemsep=2pt, parsep=0pt, leftmargin=*]
\item We identify semantic errors as a core bottleneck in LLM-based optimization modeling and design a dedicated semantic reward model for error localization and faithfulness evaluation.
\item We introduce a semantic-guided correction mechanism that uses the reward model to guide hierarchical search, and an adaptive gating mechanism balances accuracy and inference cost.
\item Extensive experiments across seven benchmarks demonstrate that \model achieves a 7.6\% accuracy improvement over the strongest baseline on complex datasets.
\end{itemize}

\section{Preliminaries and Framework}

\begin{figure}[t]
    \centering
    \includegraphics[width=\linewidth]
    {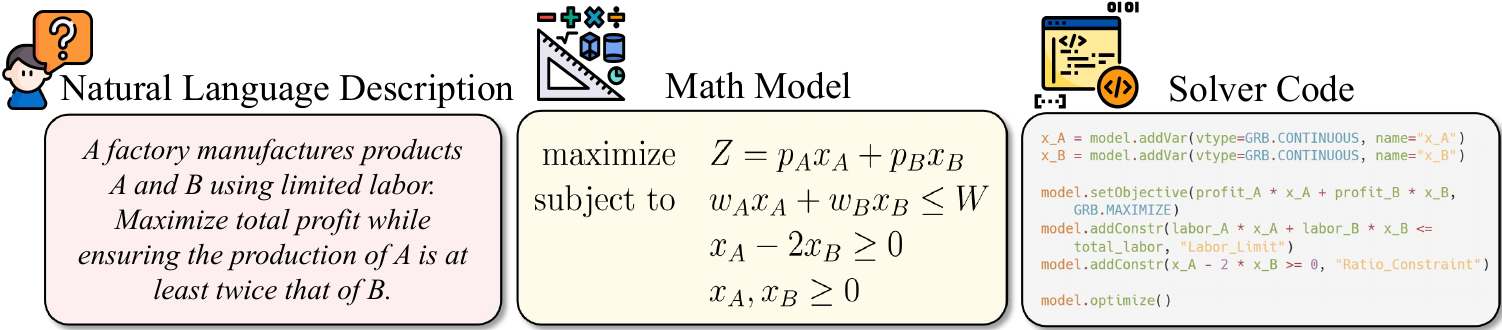}
    \caption{Overview of the three stages in optimization modeling process.}
    \label{fig:preliminaries}
\end{figure}

\subsection{Notation}
\label{sec:notation}
As illustrated in Figure~\ref{fig:preliminaries}, optimization modeling successively connects three objects: a natural-language description, a math model, and  a solver code~\cite{yangLargeLanguageModels2023}.

\paratitle{Natural Language Description.}
Let $P$ denote the input problem description. It specifies the optimization context, parameters, objective intent, decision logic, and constraints in natural language.

\paratitle{Math Model.}
A math model $M$ formalizes $P$ through decision variables $\boldsymbol{x}$, an objective $f(\boldsymbol{x})$, and constraints. Let $\boldsymbol{g}(\boldsymbol{x}) \le \mathbf{0}$ and $\boldsymbol{h}(\boldsymbol{x}) = \mathbf{0}$ denote inequality and equality constraints, respectively. Formally, $M$ is formulated as:
\begin{equation}\small
    \min_{\boldsymbol{x}} f(\boldsymbol{x}), \quad \text{s.t. } \boldsymbol{g}(\boldsymbol{x}) \le \mathbf{0}, \quad \boldsymbol{h}(\boldsymbol{x}) = \mathbf{0}.
\end{equation}
\paratitle{Solver Code.}
Let $C$ denote executable solver code that implements $M$, including the variables, objective, and constraints. Executing $C$ with a solver returns an objective value $y$ or an execution failure.

\begin{figure*}[t]
    \centering
    \includegraphics[width=\textwidth]{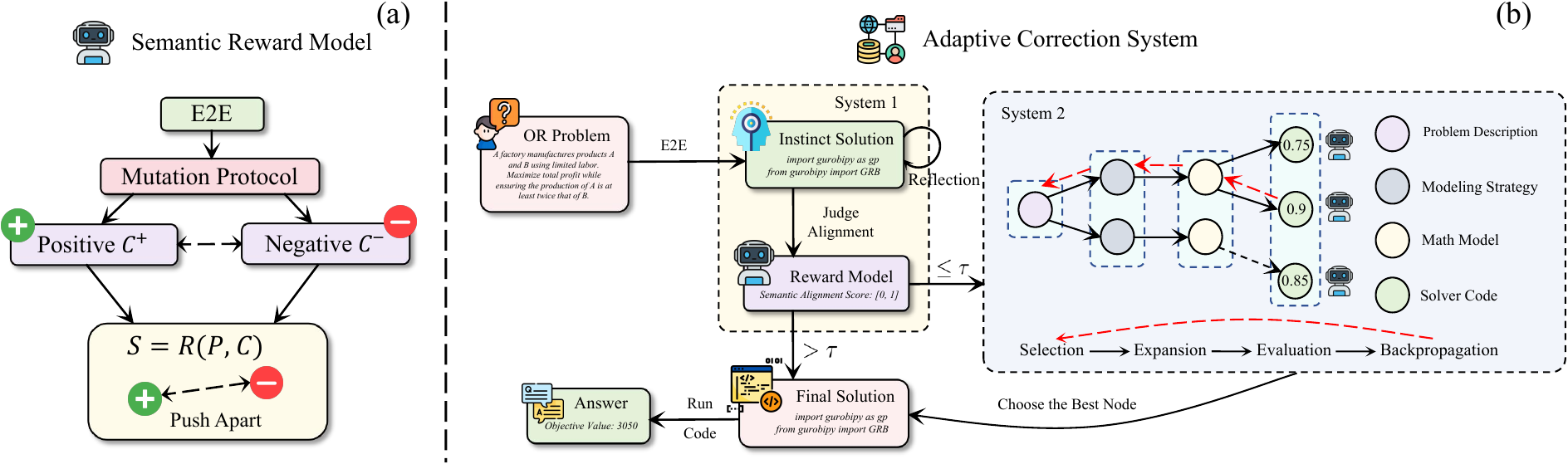}
    \caption{\textbf{Overview of the \model framework.} (a) Training pipeline of the semantic reward model. (b) Inference architecture of the adaptive correction system.}
    \label{fig:framework}
\end{figure*}

\subsection{Problem Statement}
\label{sec:pro_sat}

Automated optimization modeling asks an LLM parameterized by $\theta$ to map a problem description $P$ to solver code $C$, denoted as $C \sim p_{\theta}(\cdot | P).$
The desired output must satisfy two requirements: it should be executable, and the implemented model should be semantically faithful to $P$. We therefore distinguish two failure modes.

\paratitle{Syntactic Error.}
A syntactic error occurs when $C$ fails to run successfully or is judged infeasible by the solver. In this case, the failure is directly exposed by execution or solver feedback.

\paratitle{Semantic Error.}
A semantic error occurs when $C$ runs successfully and returns an objective value $y$, but the implemented math model is not faithful to the intent of $P$. Such errors arise from semantic-level modeling mismatches, including misunderstood constraints, incorrect objectives, or invalid modeling assumptions.

\subsection{Framework Overview}
\label{sec:overview}
As illustrated in Figure~\ref{fig:framework}, \model corrects semantic errors with a semantic reward model and an adaptive correction system. Figure~\ref{fig:framework}(a) shows how the reward model is trained to evaluate logical consistency, while Figure~\ref{fig:framework}(b) shows how the trained model guides inference-time correction.

Given a problem description $P$, \model first samples initial solver code $C_{init} \sim p_{\theta}(\cdot | P)$. It then applies a solver-informed evaluation: the code is executed, and if a syntax error occurs, the error message is appended to the context for up to $K=3$ rounds of reflection until the code runs or the budget is exhausted. The final confidence score is $S = \mathbb{I}(C) \cdot R(P, C)$, where $\mathbb{I}(C)=1$ if the code executes successfully after reflection and $0$ otherwise, and $R$ is the semantic reward model. The resulting score for the initial solver code is $S_{init} = \mathbb{I}(C_{init}) \cdot R(P, C_{init})$.

Next, \model applies an adaptive gate with threshold $\tau$. If $S_{init} > \tau$, the framework accepts $C_{init}$ and stops, avoiding unnecessary computation. If $S_{init} \le \tau$, the initial solver code falls below the semantic confidence threshold, and \model activates semantic-guided MCTS. The search decomposes correction into modeling strategy, mathematic model, and solver code, uses $R$ as the value signal to navigate the tree, and returns refined solver code $C$.

\section{Semantic Reward Model}
\label{sec:identification}

We train a semantic reward model $R$ to assess whether generated solver code is faithful to the input problem. We instantiate $R$ with a pre-trained LLM backbone~\cite{michailidisConstraintModellingLLMs2024} and replace the language modeling head with a value head that outputs a continuous scalar. During inference, this scalar is normalized with a sigmoid function to produce a confidence score in $[0,1]$ for adaptive gating. During training, the raw value is used in the contrastive objective. The training pipeline consists of two steps: constructing preference data and learning $R$ from pairwise comparisons.

\subsection{Data Construction}
We construct a preference dataset $\mathcal{D}_{pref}$ from $\mathcal{D}_{base}=\{(P_k,y_k^*)\}_{k=1}^{K}$, where $P_k$ is a problem description and $y_k^*$ is the optimal value. For each problem, we define one preference entry as
\begin{equation}\small
    \mathcal{E}_k = (P_k, \mathcal{C}_k^+, \mathcal{C}_k^-).
\end{equation}
Here, $\mathcal{C}_k^+$ contains solver code that runs successfully, is semantically faithful to $P_k$, and returns $y_k^*$, whereas $\mathcal{C}_k^-$ contains solver code that runs successfully and returns objective values but is semantically incorrect. The full dataset is
\begin{equation}\small
    \mathcal{D}_{pref} = \{\mathcal{E}_1, \mathcal{E}_2, \dots, \mathcal{E}_K\}.
\end{equation}

\paratitle{Trajectory Collection.} For each $P_k$, we sample $N_{\mathrm{cand}}$ candidate solver code instances $\{C_{k,i}\}_{i=1}^{N_{\mathrm{cand}}}$ from the generator $p_\theta(\cdot|P_k)$ and execute each candidate to obtain an objective value $\hat{y}_{k,i}$. Let $r_{k,i}=|\hat{y}_{k,i}-y_k^*|/|y_k^*|$. After discarding candidates that fail to run or are judged infeasible by the solver, we initialize the preference sets according to objective-value correctness:
\begin{equation}\small
    \begin{aligned}
    C_{k,i} \in \mathcal{C}_k^+ &\iff r_{k,i}<10^{-6},\\
    C_{k,i} \in \mathcal{C}_k^- &\iff r_{k,i}\ge 10^{-6}.
    \end{aligned}
\end{equation}
Only candidates that run successfully and return objective values are retained, so $\mathcal{C}_k^-$ captures semantic rather than syntactic failures.

\paratitle{Trajectory Augmentation.}
Natural sampling covers frequent errors but leaves long-tail semantic traps underrepresented. We therefore augment each entry $\mathcal{E}_k$ with hard negatives produced by targeted mutations~\cite{liMITSEnhancedTree2025}. For each positive code instance $C \in \mathcal{C}_k^+$, we sample $L$ mutation operators $\{m_\ell\}_{\ell=1}^{L}$ from Table~\ref{tab:mutation_operators} in Appendix~\ref{sec:mutation_operators}. Each operator is injected into $C$ independently, producing $L$ mutated code instances $\{\tilde{C}_{\ell}\}_{\ell=1}^{L}$. Candidates that fail to run or are judged infeasible by the solver are discarded, and the remaining mutants that return objective values are added to $\mathcal{C}_k^-$ as hard semantic negatives. This augmentation forces $R$ to distinguish faithful math models from structurally plausible but semantically wrong solver code.

\subsection{Contrastive Training}
The reward model $R$ is trained to measure semantic consistency between a problem description $P$ and solver code $C$. A straightforward approach is supervised fine-tuning (SFT)~\cite{SFT}, which assigns label $1$ to $C \in \mathcal{C}_k^+$ and label $0$ to $C \in \mathcal{C}_k^-$. This formulation trains on individual labeled examples, limiting the amount of supervision and making the model more prone to surface-pattern fitting rather than generalization.

We therefore train $R$ with a Bradley-Terry pairwise objective~\cite{sunRethinkingRewardModeling2024}. For each entry $\mathcal{E}_k$, we sample a winner $C_{win}$ from the correct code set $\mathcal{C}_k^+$ and a loser $C_{lose}$ from the incorrect code set $\mathcal{C}_k^-$. This pairwise construction expands the effective training set by pairing correct code instances with multiple incorrect ones, and it reduces the tendency to fit surface patterns in individual labeled examples. We define the reward margin as
\begin{equation}\small
    \Delta_\theta
    = R_\theta(P_k,C_{win}) - R_\theta(P_k,C_{lose}).
\end{equation}
The semantic reward model is optimized by minimizing the negative log-likelihood of the winner being preferred over the loser:
\begin{equation}\small
\mathcal{L}(\theta) = - \mathbb{E}_{(P_k, C_{win}, C_{lose}) \sim \mathcal{D}_{pref}}[\log \sigma(\Delta_\theta)].
\end{equation}
Here, $\sigma(\cdot)$ is the sigmoid function. This objective trains $R$ to assign higher semantic consistency scores to faithful solver code than to solver code that runs successfully and returns objective values but is semantically incorrect.

\section{Adaptive Correction System}
Using the semantic signal from $R$, we build an adaptive correction system that repairs semantic errors through hierarchical search while controlling inference cost to remain acceptable. The system balances single-pass generation and deeper reasoning with an adaptive gate~\cite{jiSurveyTestTimeCompute2025}: initial solver code with high semantic confidence is accepted directly, whereas lower-confidence code is routed to MCTS-based correction.

\subsection{Gating via Direct Inference}
Given a problem description $P$, we first sample initial solver code from the generator, denoted as $C_{init} \sim p_\theta(\cdot | P).$ Then the code is executed, reflected upon for up to $K$ rounds if syntax errors arise, and scored as $S_{init}=\mathbb{I}(C_{init}) \cdot R(P,C_{init})$. Given a threshold $\tau$, the gate compares $S_{init}$ against $\tau$ to determine the final solver code $C$:

\begin{equation}\small
    C =
    \begin{cases}
    C_{init}, & S_{init} > \tau, \\
    \mathrm{MCTS}(P), & S_{init} \leq \tau.
    \end{cases}
\end{equation}

When $S_{init} > \tau$, the system returns $C_{init}$ directly and avoids additional inference. Otherwise, the initial solver code falls below the semantic confidence threshold, and the system activates the hierarchical search described below for correction.

\subsection{Hierarchical Decomposition}
When search is triggered, we use MCTS to explore the hypothesis space~\cite{ dingDynamicParallelTree2025}. Prior search-based methods are themselves hierarchical over formulation components such as decision variables, objectives, and constraints~\cite{astorgaAutoformulationMathematicalOptimization2025}, but their search stays within a single level of modeling abstraction. In contrast, our search follows the modeling process of human OR experts. It decomposes solver code generation into a three-stage trajectory $\mathcal{S}=(v^{(1)},v^{(2)},v^{(3)})$, organized as a search tree $\mathcal{T}$ rooted at $v^{(0)}$:

\begin{itemize}[topsep=2pt, itemsep=2pt, parsep=0pt, leftmargin=*]
    \item \textbf{Modeling Strategy ($v^{(1)}$)} defines decision variables and selects the optimization framework, such as LP or MILP.
    \item \textbf{Math Model ($v^{(2)}$)} mathematically formalizes the objective and constraints.
    \item \textbf{Solver Code ($v^{(3)}$)} implements the math model as executable solver code $C$.
\end{itemize}
This hierarchy factorizes the generation policy as
\begin{equation}\small
    \pi_\theta(\mathcal{S} | P) = \prod_{k=1}^{3} \pi_\theta(v^{(k)} | v^{(<k)}, P),
\end{equation}
where $v^{(<k)}=\{v^{(0)},\dots,v^{(k-1)}\}$ denotes the partial modeling history. This decomposition enables MCTS to revise high-level modeling choices before committing to equations or code.

\subsection{MCTS Process}
The MCTS procedure, detailed in Algorithm~\ref{alg:mcts} in Appendix~\ref{sec:appendix_mcts_algorithm}, starts from the root $v^{(0)}$ of the hierarchical tree $\mathcal{T}$. Each node $v$ stores a visit count $N(v)$ and an estimated semantic value $Q(v)$, both initialized to zero. Each iteration contains four phases as follows.

\noindent\textbf{1. Selection.} Starting from $v^{(0)}$, the algorithm follows the child with the largest UCT score until reaching a leaf $\tilde{v}$:
\begin{equation}\small
    UCT(v^{(i+1)}) = Q(v^{(i+1)}) + \omega \cdot \sqrt{\frac{\ln N(v^{(i)})}{N(v^{(i+1)}) + \epsilon}},
\end{equation}
where $\omega$ controls exploration and $\epsilon$ prevents division by zero. For newly expanded nodes that have not yet received semantic rewards, we follow Autoformulator~\cite{astorgaAutoformulationMathematicalOptimization2025} and use the same LLM as the policy network to assign initial prior scores. These priors support the cold-start choice among unscored successors.

\noindent\textbf{2. Expansion.} If $\tilde{v}$ is non-terminal at depth $i$, we expand it by sampling $H$ successors from the policy network:
\begin{equation}\small
    v^{(i+1)}_j \sim \pi_\theta(\cdot | \tilde{v}, P), \quad j \in \{1, \dots, H\}.
\end{equation}
These nodes are added to $\mathcal{T}$, and the successor with the highest LLM prior is selected as $v^*$ for simulation. If $\tilde{v}$ is already terminal, we set $v^*=\tilde{v}$. The prompts for each expansion layer are provided in Appendix~\ref{sec:prompts}.

\noindent\textbf{3. Simulation.} Starting from $v^*$, the rollout recursively samples lower-level states until a terminal solver code $C$ is obtained. The system then applies the same solver-informed evaluation: $C$ is executed, and upon syntax errors, the model reflects with appended error messages for up to $K$ rounds. The final reward is computed as $S = \mathbb{I}(C) \cdot R(P, C)$, where $\mathbb{I}(C)=1$ if the code runs successfully after reflection and $0$ otherwise.

\noindent\textbf{4. Backward.} The score $S$ is propagated from $v^*$ to $v^{(0)}$. For every node $v$ on this path, we update:
\begin{equation}\small
    \begin{aligned}
    N(v) &\leftarrow N(v)+1,\\
    Q(v) &\leftarrow Q(v) + \frac{S - Q(v)}{N(v)}.
    \end{aligned}
\end{equation}
The process repeats iteratively until the budget $T_{max}$ is exhausted.


In this way, MCTS repeatedly rolls out partial modeling states to terminal solver code and transfers semantic consistency scores back to intermediate nodes. The resulting $Q(v)$ values estimate how semantically promising each modeling branch is, therefore enabling the search to suppress flawed intermediate choices and converge efficiently toward a correct solver code.

\section{Experiments}

We evaluate \model through the following three research questions:
\begin{itemize}[topsep=2pt, itemsep=2pt, parsep=0pt, leftmargin=*]
    \item \textbf{RQ1: Overall Performance.} How does \model compare with baselines across benchmarks?
    \item \textbf{RQ2: Component Analysis.} Are the proposed modules effective, and does \model generalize across LLM backbones?
    \item \textbf{RQ3: Efficiency.} How does \model trade off accuracy and computational cost?
\end{itemize}

\begin{table*}[t]
\centering
\caption{Main results on seven optimization benchmarks. The best results are highlighted in \textbf{bold}, and the second-best results are \underline{underlined}. Red arrows indicate absolute performance gains over the strongest baseline.}
\label{tab:main_results}
\resizebox{\textwidth}{!}{%
\begin{tabular}{llccccccc}
\toprule
\multirow{2}{*}{\textbf{Category}} & \multirow{2}{*}{\textbf{Method}} & \multicolumn{4}{c}{\textbf{Standard Datasets}} & \multicolumn{3}{c}{\textbf{Complex Datasets}} \\
\cmidrule(lr){3-6}\cmidrule(lr){7-9}
& & \textbf{NL4Opt} & \textbf{NL4LP} & \textbf{EasyLP} & \textbf{ReSocratic} & \textbf{IndustryOR} & \textbf{ComplexLP} & \textbf{ComplexOR} \\ \midrule
\multirow{2}{*}{Standard} 
 & Prompting & 65.4\% & 75.8\% & 87.2\% & 67.5\% & 47.6\% & 41.4\% & 44.4\% \\
 & Reflection & 65.0\% & 76.4\% & 87.2\% & 68.0\% & 54.8\% & 39.6\% & 38.9\% \\ \midrule
\multirow{4}{*}{Workflow} 
 & OptiMUS & 79.0\% & 90.5\% & 93.6\% & 78.7\% & 57.1\% & 45.1\% & 59.3\% \\
 & Chain-of-Experts & 74.8\% & 90.5\% & 93.2\% & 79.4\% & 59.5\% & 46.0\% & 50.0\% \\
 & OptiTree & 73.8\% & 82.0\% & 93.4\% & 72.5\% & 47.6\% & 73.8\% & 63.0\% \\
 & SAC-Opt & 85.1\% & 95.5\% & 94.3\% & 89.3\% & 57.1\% & 74.8\% & 55.6\% \\ \midrule
\multirow{2}{*}{Fine-tuned} 
 & ORLM & 73.8\% & 76.4\% & 90.4\% & 61.8\% & 42.9\% & 59.5\% & 50.0\% \\
 & SIRL & 86.4\% & 95.5\% & \underline{98.1\%} & 87.8\% & 45.2\% & 80.2\% & 44.4\% \\ \midrule
\multirow{2}{*}{Search-based} 
 & Autoformulator & 76.2\% & 85.4\% & 94.9\% & 79.4\% & 49.2\% & 52.3\% & 48.1\% \\
 & SolverLLM & 82.7\% & 88.8\% & 94.5\% & 83.1\% & 50.8\% & 68.5\% & 57.4\% \\ \midrule
\midrule
\multirow{2}{*}{\textbf{Ours}} 
  & {\textbf{\model (BoN)}} & \underline{87.9}\% \textcolor{red}{\scriptsize ($\uparrow$1.5\%)} & \underline{96.1\%} \textcolor{red}{\scriptsize ($\uparrow$0.6\%)} & 97.6\% & \underline{89.8\%} \textcolor{red}{\scriptsize ($\uparrow$0.5\%)} & \underline{64.3\%} \textcolor{red}{\scriptsize ($\uparrow$4.8\%)} & \underline{82.9\%} \textcolor{red}{\scriptsize ($\uparrow$2.7\%)} & \underline{66.7\%} \textcolor{red}{\scriptsize ($\uparrow$3.7\%)} \\
 & {\textbf{\model (P@N)}} & \textbf{91.1\%} \textcolor{red}{\scriptsize ($\uparrow$4.7\%)} & \textbf{99.4\%} \textcolor{red}{\scriptsize ($\uparrow$3.9\%)} & \textbf{98.5\%} \textcolor{red}{\scriptsize ($\uparrow$0.4\%)} & \textbf{93.1\%} \textcolor{red}{\scriptsize ($\uparrow$3.8\%)} & \textbf{66.7\%} \textcolor{red}{\scriptsize ($\uparrow$7.2\%)} & \textbf{86.5\%} \textcolor{red}{\scriptsize ($\uparrow$6.3\%)} & \textbf{72.2\%} \textcolor{red}{\scriptsize ($\uparrow$9.2\%)} \\ \bottomrule
\end{tabular}%
}
\end{table*}

\subsection{Experimental Setup}

\paratitle{Datasets.} We evaluate on seven benchmarks~\cite{xiaoSurveyOptimizationModeling2025} divided into two groups: (1) \emph{Standard Datasets}: NL4Opt~\cite{ramamonjisonNL4OptCompetitionFormulating2023}, NL4LP~\cite{ahmaditeshniziOptiMUSScalableOptimization2024}, EasyLP~\cite{huangLLMsMathematicalModeling2025}, and ReSocratic~\cite{yangOptiBenchMeetsReSocratic2024}, which primarily contain standard linear programming problems; and (2) \emph{Complex Datasets}: IndustryOR~\cite{huangORLMCustomizableFramework2025}, ComplexLP~\cite{huangLLMsMathematicalModeling2025}, and ComplexOR~\cite{xiaoChainofExpertsWhenLLMs2023}, which include implicit constraints, long-context descriptions and serve as the primary testbed for semantic alignment.

\paratitle{Baselines.} We compare \model with ten baselines in four groups: (1) \emph{Standard}: Standard Prompting and Self-Reflection~\cite{reflection}; (2) \emph{Workflow}: OptiMUS~\cite{ahmaditeshniziOptiMUS03UsingLarge2025}, Chain-of-Experts~\cite{xiaoChainofExpertsWhenLLMs2023}, OptiTree~\cite{liuOptiTreeHierarchicalThoughts2025}, and SAC-Opt~\cite{zhangOptimizationModelingSemantic2025}; (3) \emph{Fine-tuned}: ORLM~\cite{huangORLMCustomizableFramework2025} and SIRL~\cite{chenSolverInformedRLGrounding2025}; (4) \emph{Search-based}: Autoformulator~\cite{astorgaAutoformulationMathematicalOptimization2025} and SolverLLM~\cite{liSolverLLMLeveragingTestTime2025}. For fairness, all baselines use the same policy model and search settings as \model, except the fine-tuned baselines, which use their own fine-tuned models.

\paratitle{Implementation Details.} We instantiate the policy networks ($p_\theta, \pi_\theta$) with \textbf{GPT-4.1 Nano} and fine-tune the semantic reward model $R$ from \textbf{Qwen3-4B-Instruct-2507}~\cite{qwen3technicalreport}. We report \textit{Accuracy}, counting a solution as correct when its solver result differs from the ground truth by less than $10^{-6}$. We also report two sampling metrics: \textit{Pass@N (P@N)}, the probability that at least one of $N$ samples is correct, and \textit{Best-of-N (BoN)}, the accuracy of the best solution selected by the reward model. Appendix~\ref{sec:appendix_implementation_details} provides the remaining implementation details.

\subsection{Overall Performance}

\paratitle{Main Results.} 
As shown in Table~\ref{tab:main_results}, \model obtains the best results on most datasets, with especially clear gains on complex benchmarks. On ComplexOR, \model reaches 72.2\% accuracy, surpassing the strongest baseline by 9.2\%. The small gap between the \textit{Best-of-N} and \textit{Pass@N} results of \model shows that the reward model reliably selects semantically faithful candidates when they appear in the sampled set.

We make further observations as follows: 1) \emph{Standard and workflow methods} lag behind on complex applied datasets, showing that generation or workflow decomposition alone is insufficient without a semantic verification signal. 2) \emph{Fine-tuned methods}, especially SIRL, are strong on simpler datasets but less consistent on complex tasks, indicating that fine-tuning helps translate explicit requirements but still misses implicit constraints in long-context problems. 3) \emph{Search-based methods} improve over single-pass generation but remain below \model. Their reliance on general-purpose evaluators and on flat search makes recovery from strategy-level modeling errors difficult, whereas \model{} backtracks across modeling strategy, math model, and code.

Objective-value matching is the standard protocol on these benchmarks, but it can credit a semantically incorrect formulation whenever that formulation happens to attain the same optimum. We therefore additionally validate \model at the formulation level on 100 LP problems with reference formulations, scoring predictions by graph edit distance and human audit alongside objective matching. \model improves over the strongest fine-tuned baseline under all three criteria, indicating that the gains extend to formulation faithfulness rather than final-answer matching alone. Appendix~\ref{sec:appendix_semantic_validation} details the protocol and the full results.

\paratitle{Robustness Analysis.}
We next recalibrate difficulty labels. Existing labels often depend on the number of constraints rather than the difficulty of finding a correct solution, yielding weak correlation with actual solver accuracy~\cite{xiaoSurveyOptimizationModeling2025}. We therefore sample three solutions from the direct generator $p_\theta(C|\mathcal{P})$ and label an instance as \textit{Easy} if all three samples are correct, \textit{Hard} if none are correct, and \textit{Medium} otherwise. This definition aligns difficulty with single-pass generation success.

Table~\ref{tab:main_results} shows that SIRL is the best or near-best baseline on multiple datasets, so we use it as the strongest baseline for the robustness comparison. The results in Table~\ref{tab:difficulty_analysis} show that \model achieves clear improvements over SIRL on complex datasets. Except for the \textit{Medium} subset of ComplexOR, which contains only a small number of instances, we can also notice that the improvement generally becomes larger as difficulty increases. This pattern supports the role of semantic-guided search in correcting hard cases where implicit requirements and high-level modeling decisions often cause errors.

\begin{table}[t]
    \centering
    \caption{Accuracy (Best-of-N) across difficulty levels. We compare \model against the single strongest baseline (SIRL) on Easy, Medium, and Hard subsets.}
    \label{tab:difficulty_analysis}
    \resizebox{0.8\linewidth}{!}{%
    \begin{tabular}{lccc}
        \toprule
        \textbf{Dataset} & \textbf{Difficulty} & \textbf{SIRL} & \textbf{\model} \\ \midrule
        \multirow{3}{*}{\textbf{NL4Opt}} 
         & Easy & 93.6\% & \textbf{94.4\%} \textcolor{red}{\scriptsize ($\uparrow$0.8\%)} \\
         & Medium & 82.0\% & \textbf{84.0\%} \textcolor{red}{\scriptsize ($\uparrow$2.0\%)} \\
         & Hard & 69.2\% & \textbf{71.8\%} \textcolor{red}{\scriptsize ($\uparrow$2.6\%)} \\ \midrule
        \multirow{3}{*}{\textbf{ComplexLP}} 
         & Easy & 88.6\% & 88.6\% \textcolor{gray}{\scriptsize (+0.0\%)} \\
         & Medium & 94.1\% & \textbf{97.1\%} \textcolor{red}{\scriptsize ($\uparrow$3.0\%)} \\
         & Hard & 54.5\% & \textbf{63.6\%} \textcolor{red}{\scriptsize ($\uparrow$9.1\%)} \\ \midrule
        \multirow{3}{*}{\textbf{ComplexOR}} 
         & Easy & 55.6\% & \textbf{77.8\%} \textcolor{red}{\scriptsize ($\uparrow$22.2\%)} \\
         & Medium & 50.0\% & 50.0\% \textcolor{gray}{\scriptsize (+0.0\%)} \\
         & Hard & 28.6\% & \textbf{57.1\%} \textcolor{red}{\scriptsize ($\uparrow$28.5\%)} \\ \bottomrule
    \end{tabular}%
    }
\end{table}

\subsection{Component Analysis}

We evaluate three variants to isolate the contribution of each component:
(1) \emph{Backbone Variants}, which replace the generation backbone with Gemini-2.5-Flash-Lite, GPT-4.1 mini, and Qwen3-Max, testing whether the correction framework generalizes across LLM backbones;
(2) \emph{Critic Variants}, which replace the fine-tuned semantic reward model with either the generation backbone (GPT-4.1 Nano) or the reward-model backbone (Qwen3-4B-Instruct-2507) during MCTS evaluation, testing whether the trained reward model provides a stronger semantic value signal;
(3) \emph{Adaptive Gating Ablation}, which removes the confidence threshold and runs MCTS for all queries, testing whether adaptive gating reduces unnecessary inference cost.

\paratitle{Backbone Variants.} Table~\ref{tab:backbone} presents the performance gains across diverse LLM backbones.
\model consistently outperforms standard prompting in \textit{Best-of-N} settings across all tested generators. These gains indicate that the semantic correction framework generalizes across LLM backbones rather than depending on a single generator.

Because the fine-tuned baselines use their own specialized generators whereas \model uses GPT-4.1 Nano, we further control for the generator itself. Keeping the semantic reward model, the adaptive gate, and the MCTS procedure fixed, we replace only \model's generator with the fine-tuned models of ORLM and SIRL. As reported in Table~\ref{tab:generator_control}, \model outperforms ORLM and SIRL on all three datasets. The improvements reflect the correction framework rather than the choice of generator.

\paratitle{Critic Variants.}
We examine the role of the semantic reward model from two angles: discrimination quality and downstream search accuracy. For discrimination quality, we evaluate whether the reward model distinguishes semantically faithful solver code from incorrect executable solver code. The test set is constructed from single-pass generation on NL4Opt, ComplexLP, and ComplexOR. As shown in Table~\ref{tab:rm_discrimination}, the fine-tuned reward model achieves the highest AUC/F1 on all three datasets compared with both its unfine-tuned backbone and the policy LLM used for generation.

\begin{table}[t]
    \centering
    \caption{Performance across LLM backbones. We compare \model with standard prompting in BoN settings.}
    \label{tab:backbone}
    \resizebox{\linewidth}{!}{%
    \begin{tabular}{clccc}
        \toprule
        \textbf{Backbone LLM} & \textbf{Method} & \textbf{NL4Opt} & \textbf{ComplexLP} & \textbf{ComplexOR} \\
        \midrule
        \multirow{2}{*}{
        \makecell{\textbf{Gemini-2.5}\\\textbf{Flash-Lite}}}
        & Prompting (BoN) & 70.1\% & 64.0\% & 55.6\% \\
        & \textbf{\model (BoN)} & \textbf{79.0\%} & \textbf{75.7\%} & \textbf{63.0\%} \\
        \midrule
        \multirow{2}{*}{\textbf{GPT-4.1 mini}}
        & Prompting (BoN) & 74.8\% & 64.8\% & 55.6\% \\
        & \textbf{\model (BoN)} & \textbf{86.9\%} & \textbf{79.3\%} & \textbf{66.7\%} \\
        \midrule
        \multirow{2}{*}{\textbf{Qwen3-Max}} 
        & Prompting (BoN) & 76.2\% & 66.7\% & 55.6\% \\
        & \textbf{\model (BoN)} & \textbf{88.8\%} & \textbf{82.0\%} & \textbf{66.7\%} \\
        \bottomrule
    \end{tabular}%
    }
\end{table}

\begin{table}[t]
    \centering
    \caption{Fine-tuned generator control on complex datasets. We replace only \model's generator with the fine-tuned models of ORLM and SIRL.}
    \label{tab:generator_control}
    \resizebox{\linewidth}{!}{%
    \begin{tabular}{lccc}
        \toprule
        \textbf{Variant} & \textbf{IndustryOR} & \textbf{ComplexLP} & \textbf{ComplexOR} \\
        \midrule
        ORLM & 42.9\% & 59.5\% & 50.0\% \\
        \model w/ ORLM generator & \textbf{61.9\%} & \textbf{74.8\%} & \textbf{66.7\%} \\
        \midrule
        SIRL & 45.2\% & 80.2\% & 44.4\% \\
        \model w/ SIRL generator & \textbf{66.7\%} & \textbf{85.6\%} & \textbf{66.7\%} \\
        \bottomrule
    \end{tabular}%
    }
\end{table}

\begin{figure}[t]
    \centering
    \includegraphics[width=\linewidth]{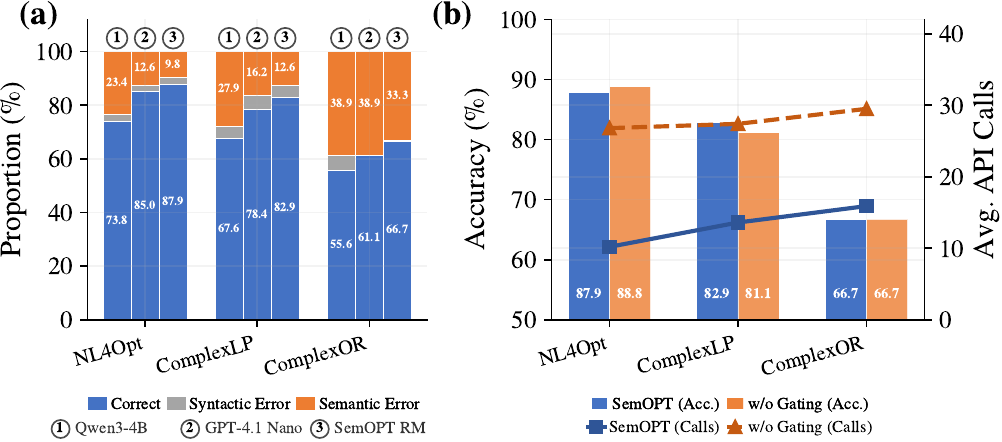}
    \caption{(a) Breakdown of model outputs under different critic models. (b) Efficiency evaluation of the adaptive gating mechanism.}
    \label{fig:overall_analysis}
\end{figure}

For downstream search accuracy, Figure~\ref{fig:overall_analysis}(a) breaks down model outputs into correct solutions, syntactic errors, and semantic errors under different critic models.
The fine-tuned semantic reward model not only achieves the highest proportion of correct solutions across all benchmarks, but also yields a lower share of semantic errors compared to both the unfine-tuned base model and general-purpose LLM. This confirms that the reward model provides more precise semantic signals, guiding the search toward semantically consistent solutions.

Because \model differs from prior search-based methods in both evaluation and search structure, Table~\ref{tab:hierarchy_ablation} separates their effects. Both controls use \model's semantic reward model. \emph{RM-only reranking} samples the same number of end-to-end solver-code candidates and selects one without search, whereas \emph{Autoformulator-style + Semantic RM} applies MCTS to formulation components such as variables, objectives, and constraints, without \model's strategy--math-model--code hierarchy. RM-only reranking performs worst, showing that reward-guided selection alone is insufficient. The Autoformulator-style variant narrows the gap, but \model remains better, supporting the value of the correction hierarchy. Conversely, replacing the reward model with a general LLM critic while keeping the \model hierarchy reduces performance by 4.8, 4.5, and 5.6 points. The evaluator and hierarchy therefore provide complementary gains.

\paratitle{Adaptive Gating Ablation.} Figure~\ref{fig:overall_analysis}(b) analyzes the trade-off between accuracy and inference cost.
This ablation removes the gate and runs MCTS for every query. This change increases API calls substantially: on NL4Opt, \textit{w/o Gating} uses about $2.6\times$ more calls than the full framework while improving accuracy by less than 1\%. Average API calls also increase as dataset difficulty increases from NL4Opt to ComplexOR, indicating that the gate allocates more search budget to harder tasks.

\begin{table}[t]
    \centering
    \caption{Reward-model discrimination performance. Each cell reports AUC / F1.}
    \label{tab:rm_discrimination}
    \resizebox{0.9\linewidth}{!}{%
    \begin{tabular}{lccc}
        \toprule
        \textbf{Dataset} & \makecell{\textbf{Qwen3-4B}\\\textbf{Instruct-2507}} & \makecell{\textbf{GPT-4.1}\\\textbf{Nano}} & \makecell{\textbf{\model}\\\textbf{RM}} \\
        \midrule
        NL4Opt & 0.546 / 0.781 & 0.596 / 0.776 & \textbf{0.744 / 0.849} \\
        ComplexLP & 0.619 / 0.760 & 0.701 / 0.785 & \textbf{0.820 / 0.800} \\
        ComplexOR & 0.576 / 0.667 & 0.611 / 0.696 & \textbf{0.833 / 0.778} \\
        \bottomrule
    \end{tabular}%
    }
\end{table}

\begin{table}[t]
    \centering
    \caption{Factorizing the semantic evaluator and the correction hierarchy on complex datasets (BoN accuracy).}
    \label{tab:hierarchy_ablation}
    \resizebox{\linewidth}{!}{%
    \begin{tabular}{lccc}
        \toprule
        \textbf{Variant} & \textbf{IndustryOR} & \textbf{ComplexLP} & \textbf{ComplexOR} \\
        \midrule
        Autoformulator & 49.2\% & 52.3\% & 48.1\% \\
        RM-only reranking & 52.4\% & 50.5\% & 44.4\% \\
        Autoformulator-style + Semantic RM & 61.9\% & 68.5\% & 61.1\% \\
        \model hierarchy + general LLM & 59.5\% & 78.4\% & 61.1\% \\
        \textbf{\model} & \textbf{64.3\%} & \textbf{82.9\%} & \textbf{66.7\%} \\
        \bottomrule
    \end{tabular}%
    }
\end{table}

\subsection{Efficiency}

We evaluate efficiency from two perspectives: whether \model improves over common iterative and search baselines under comparable budgets, and whether reward-guided search avoids the cost of exhaustive hypothesis enumeration.

\paratitle{Comparison with Iterative and Search Baselines.}
Table~\ref{tab:efficiency_iterative_search} compares \model with two classic iterative and search baselines, Self-Refinement (SR) and Beam Search (BS), on the combined hard subsets of NL4Opt, ComplexLP, and ComplexOR. To evaluate the efficiency of \model itself, we set the iteration count of SR and the beam-search budget of BS so that both baselines use no less inference cost than \model. Under this setting, \model achieves the highest accuracy while using less time and fewer API calls. This result shows that the improvement of \model comes from more effective semantic reward-guided correction rather than a larger inference budget.

\begin{table}[t]
    \centering
    \caption{Efficiency comparison with iterative and search baselines on the combined hard subsets of NL4Opt, ComplexLP, and ComplexOR.}
    \label{tab:efficiency_iterative_search}
    \resizebox{0.8\linewidth}{!}{%
    \begin{tabular}{lccc}
        \toprule
        \textbf{Method} & \textbf{Acc.} & \textbf{Time (s)} & \textbf{API Calls} \\
        \midrule
        SR (25 iters) & 55.7\% & 61.2 & 26.0 \\
        BS (beam 2, branch 5) & 63.3\% & 38.9 & 30.0 \\
        \textbf{\model} & \textbf{67.1}\% & \textbf{32.3} & \textbf{25.1} \\
        \bottomrule
    \end{tabular}%
    }
\end{table}

\paratitle{Comparison with Exhaustive Search.}
We next compare \model with a brute-force Naive Search (NS) baseline that exhaustively evaluates a fixed candidate space. As shown in Table~\ref{tab:efficiency_naive_search}, NS obtains slightly higher accuracy on NL4Opt and ComplexLP, but requires 84 API calls per instance. \model matches NS on ComplexOR and trails it by only 0.9 average accuracy points across the three datasets, while reducing API calls to 12--19\% of NS. These results show that \model preserves nearly the same accuracy as exhaustive search but avoids most of its inference cost.

\begin{table}[t]
    \centering
    \caption{Comparison with exhaustive Naive Search (NS). Each cell reports accuracy / API calls.}
    \label{tab:efficiency_naive_search}
    \resizebox{0.8\linewidth}{!}{%
    \begin{tabular}{lccc}
        \toprule
        \textbf{Method} & \textbf{NL4Opt} & \textbf{ComplexLP} & \textbf{ComplexOR} \\
        \midrule
        NS & 89.7\% / 84.0 & 83.8\% / 84.0 & 66.7\% / 84.0 \\
        \textbf{\model} & 87.9\% / 10.2 & 82.9\% / 13.6 & 66.7\% / 15.9 \\
        \bottomrule
    \end{tabular}%
    }
\end{table}

\begin{figure}[t]
    \centering
    \includegraphics[width=\linewidth]{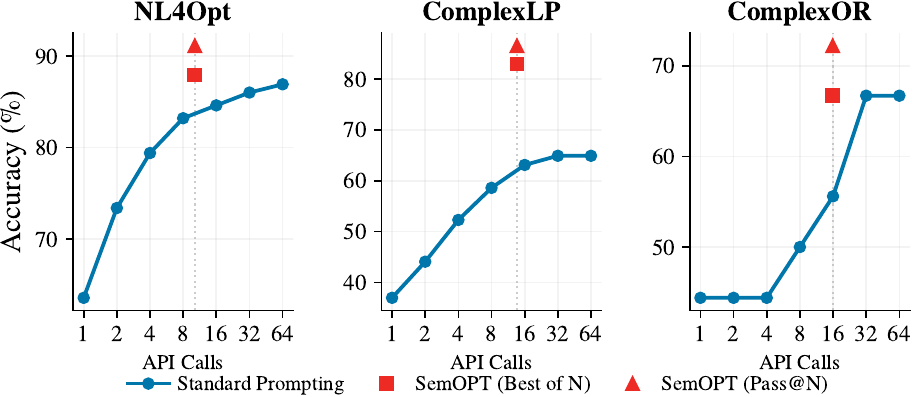}
    \vspace{-10pt}
    \caption{Comparison between \model and the standard prompting baseline across optimization benchmarks.}
    \label{fig:pass_at_n}
    \vspace{-10pt}
\end{figure}

\paratitle{Comparison with Pass@N Scaling.}
Figure~\ref{fig:pass_at_n} further compares \model with standard prompting under Pass@N scaling. \model lies on the upper-left side of the prompting curves, achieving higher accuracy with fewer API calls. For example, on ComplexLP, \model reaches 86.5\% accuracy with 13.6 calls on average, whereas standard prompting saturates at 64.9\% even with 64 calls. These results indicate that semantic guidance changes the cost-accuracy trade-off rather than merely increasing the number of sampled code instances.

\section{Related Work}

\paratitle{LLMs for Optimization Modeling.} Recent work on LLM-based optimization modeling falls into four groups. Standard methods such as Reflexion~\cite{reflection} refine math models through iterative feedback. Workflow frameworks, including OptiMUS~\cite{ahmaditeshniziOptiMUS03UsingLarge2025} and Chain-of-Experts~\cite{xiaoChainofExpertsWhenLLMs2023}, decompose modeling into multi-agent pipelines. Fine-tuned models, such as ORLM~\cite{huangORLMCustomizableFramework2025} and SIRL~\cite{chenSolverInformedRLGrounding2025}, adapt LLMs with domain-specific instruction tuning or reinforcement learning. Search-based approaches, including Autoformulator~\cite{astorgaAutoformulationMathematicalOptimization2025} and SolverLLM~\cite{liSolverLLMLeveragingTestTime2025}, explore broader hypothesis spaces beyond single-pass generation. These methods improve modeling ability, but mainly target feasible or superficially plausible math models rather than semantic consistency with problem requirements.

\paratitle{Error Correction in Optimization Modeling.} LLMs produce modeling errors through hallucinated or misinterpreted problem structure~\cite{xieMURKAMultiRewardReinforcement2025}, commonly categorized as syntactic or semantic errors~\cite{wangSemGuardRealTimeSemantic2025}. Syntactic errors occur when generated solver code fails to run or is judged infeasible; execution or solver feedback directly exposes them. OptLLM~\cite{zhangSolvingGeneralNaturalLanguageDescription2024} and ORThought~\cite{yangAutomatedOptimizationModeling2025} refine code from tracebacks, while LLMOPT~\cite{jiangLLMOPTLearningDefine2024} reduces such errors through domain-specific fine-tuning. Semantic errors occur when solver code runs and returns an objective value but violates the original problem intent through misunderstood constraints, incorrect objectives, or invalid modeling assumptions. Although semantic error correction has been studied for LLM-generated content~\cite{yang2024supercorrect, guSemanticbasedOptimizationApproach2025, islam2024enhancing}, work targeting optimization modeling remains limited. Autoformulator~\cite{astorgaAutoformulationMathematicalOptimization2025} and SolverLLM~\cite{liSolverLLMLeveragingTestTime2025} address this issue through broader search and candidate selection, but rely on general-purpose verifiers that lack the domain sensitivity needed to separate plausible math models from semantically correct ones.

\section{Conclusion}
We propose \model, a framework for correcting semantic errors in LLM-based optimization modeling. \model combines a pairwise-trained semantic reward model with semantic-guided MCTS over modeling strategies, math models, and solver code, while adaptive gating invokes search only when the initial solver code falls below the semantic confidence threshold. Comprehensive experiments across seven benchmarks show that \model achieves the best results on most datasets, with especially strong gains on complex benchmarks. These results support semantic-guided search for reliable optimization modeling.

\section*{Limitations}

This work focuses on automated optimization modeling from natural-language problem descriptions to executable solver code. Although \model improves semantic consistency across seven benchmarks, our evaluation is still centered on benchmark tasks rather than deployed decision-making workflows. The framework also assumes access to solver execution and reward-model training data for constructing semantic supervision. Future work will further investigate this direction in broader real-world settings, including domain transfer, more diverse optimization tasks, and more efficient semantic-guided search.

\section*{Ethical Considerations}

\model is intended as a research framework for improving the reliability of LLM-based optimization modeling. The main societal risk is over-reliance on automatically generated solver code in high-stakes decision-making settings. Executable optimization models still sometimes encode incorrect objectives, omit important constraints, or reflect inappropriate assumptions about the problem context. Such errors create risks of harmful decisions when used without expert review. We therefore recommend that generated models be audited by domain experts before deployment. The work uses existing optimization-modeling benchmarks and does not collect human-subject data. The framework also uses commercial and open-weight LLMs, so reproducibility and environmental cost depend on model availability, API behavior, and compute budget. AI assistance was used only for language polishing and basic code drafting; it was not used to formulate the research idea, design the experiments, derive the claims, or generate the results.

\section*{Acknowledgements}

Prof. Jingyuan Wang's work is supported by the National Natural Science Foundation of China (No. 72242101), the Science and Technology Development Fund Macau SAR (0052/2023/RIA1), and State Key Laboratory of Complex \& Critical Software Environment (SKLCCSE-2025ZX-17). Prof. Zizhuo Wang's research is partially supported by the National Natural Science Foundation of China (NSFC) [Grant 72394361, 72425013], the Guangdong Provincial Key Laboratory of Mathematical Foundations for Artificial Intelligence (2023B1212010001), and the 1+1+1 CUHK-CUHK(SZ)-GDSTC Joint Collaboration Fund No 2025A0505000079.

\bibliography{5-formal}

@article{ahmaditeshniziOptiMUS03UsingLarge2025,
  title={OptiMUS-0.3: Using Large Language Models to Model and Solve Optimization Problems at Scale}, 
  author = {AhmadiTeshnizi, Ali and Gao, Wenzhi and Brunborg, Herman and Talaei, Shayan and Lawless, Connor and Udell, Madeleine},
  journal = {arXiv:2407.19633},
  year = {2024}
}

@inproceedings{ahmaditeshniziOptiMUSScalableOptimization2024,
  title = {{O}pti{MUS}: Scalable Optimization Modeling with ({MI}){LP} Solvers and Large Language Models},
  booktitle = {Proc. of ICML},
  author = {Ahmaditeshnizi, Ali and Gao, Wenzhi and Udell, Madeleine},
  year = {2024},
  pages = {577--596}
}

@inproceedings{astorgaAutoformulationMathematicalOptimization2025,
  title = {Autoformulation of Mathematical Optimization Models Using {LLM}s},
  booktitle = {Proc. of ICML},
  author = {Astorga, Nicol{\'a}s and Liu, Tennison and Xiao, Yuanzhang and van der Schaar, Mihaela},
  year = {2025},
  pages = {1864--1886}
}

@inproceedings{chenSolverInformedRLGrounding2025,
  title = {Solver-Informed RL: Grounding Large Language Models for Authentic Optimization Modeling},
  booktitle = {Proc. of NeurIPS},
  author = {Chen, Yitian and Xia, Jingfan and Shao, Siyu and Ge, Dongdong and Ye, Yinyu},
  year = {2025},
  pages = {106027--106069}
}

@inproceedings{dingDynamicParallelTree2025,
  title = {Dynamic Parallel Tree Search for Efficient {LLM} Reasoning},
  booktitle = {Proc. of ACL},
  author = {Ding, Yifu and Jiang, Wentao and Liu, Shunyu and Jing, Yongcheng and Guo, Jinyang and Wang, Yingjie and Zhang, Jing and Wang, Zengmao and Liu, Ziwei and Du, Bo and Liu, Xianglong and Tao, Dacheng},
  year = {2025},
  pages = {11233--11252}
}

@inproceedings{huangLLMsMathematicalModeling2025,
  title = {{LLM}s for Mathematical Modeling: Towards Bridging the Gap between Natural and Mathematical Languages},
  booktitle = {Proc. of NAACL Findings},
  author = {Huang, Xuhan and Shen, Qingning and Hu, Yan and Gao, Anningzhe and Wang, Benyou},
  year = {2025},
  pages = {2678--2710}
}

@article{huangORLMCustomizableFramework2025,
  title = {{ORLM}: A Customizable Framework in Training Large Models for Automated Optimization Modeling},
  journal = {Operations Research},
  author = {Huang, Chenyu and Tang, Zhengyang and Hu, Shixi and Jiang, Ruoqing and Zheng, Xin and Ge, Dongdong and Wang, Benyou and Wang, Zizhuo},
  year = {2025},
  volume = {73},
  number = {6},
  pages = {2986--3009}
}

@inproceedings{jiangLLMOPTLearningDefine2024,
  title = {LLMOPT: Learning to Define and Solve General Optimization Problems from Scratch},
  booktitle = {Proc. of ICLR},
  author = {Jiang, Caigao and Shu, Xiang and Qian, Hong and Lu, Xingyu and Zhou, Jun and Zhou, Aimin and Yu, Yang},
  year = {2025}
}

@article{jiSurveyTestTimeCompute2025,
  title = {A Survey of Test-Time Compute: From Intuitive Inference to Deliberate Reasoning},
  journal = {Computational Linguistics},
  author = {Ji, Yixin and Li, Juntao and Xiang, Yang and Ye, Hai and Wu, Kaixin and Yao, Kai and Xu, Jia and Mo, Linjian and Zhang, Min},
  year = {2026},
  pages = {1--51}
}

@inproceedings{liMITSEnhancedTree2025,
  title = {MITS: Enhanced Tree Search Reasoning for LLMs via Pointwise Mutual Information},
  booktitle = {Proc. of PAKDD},
  author = {Li, Jiaxi and Shi, Yucheng and Huang, Xiao and Lu, Jin and Liu, Ninghao},
  year = {2026},
  pages = {288--300}
}

@inproceedings{liSolverLLMLeveragingTestTime2025,
  title = {SolverLLM: Leveraging Test-Time Scaling for Optimization Problem via LLM-Guided Search},
  booktitle = {Proc. of NeurIPS},
  author = {Li, Dong and Zhao, Xujiang and Yu, Linlin and Liu, Yanchi and Cheng, Wei and Chen, Zhengzhang and Chen, Zhong and Chen, Feng and Zhao, Chen and Chen, Haifeng},
  year = {2025},
  pages = {100028--100058}
}

@inproceedings{liuOptiTreeHierarchicalThoughts2025,
  title = {OptiTree: Hierarchical Thoughts Generation with Tree Search for LLM Optimization Modeling},
  booktitle = {Proc. of NeurIPS},
  author = {Liu, Haoyang and Wang, Jie and Cai, Yuyang and Han, Xiongwei and Kuang, Yufei and Hao, Jianye},
  year = {2025},
  pages = {120713--120781}
}

@inproceedings{michailidisConstraintModellingLLMs2024,
  title = {Constraint Modelling with LLMs Using In-Context Learning},
  booktitle = {30th {{International Conference}} on {{Principles}} and {{Practice}} of {{Constraint Programming}}},
  author = {Michailidis, Kostis and Tsouros, Dimos and Guns, Tias},
  year = {2024},
  pages = {20:1--20:27},
}

@inproceedings{ramamonjisonNL4OptCompetitionFormulating2023,
  title = {NL4Opt Competition: Formulating Optimization Problems Based on Their Natural Language Descriptions},
  booktitle = {Proc. of the NeurIPS 2022 Competitions Track},
  author = {Ramamonjison, Rindranirina and Yu, Timothy and Li, Raymond and Li, Haley and Carenini, Giuseppe and Ghaddar, Bissan and He, Shiqi and Mostajabdaveh, Mahdi and Banitalebi-Dehkordi, Amin and Zhou, Zirui and Zhang, Yong},
  year = {2022},
  pages = {189--203}
}

@inproceedings{xiaoChainofExpertsWhenLLMs2023,
  title = {Chain-of-Experts: When LLMs Meet Complex Operations Research Problems},
  booktitle = {Proc. of ICLR},
  author = {Xiao, Ziyang and Zhang, Dongxiang and Wu, Yangjun and Xu, Lilin and Wang, Yuan and Han, Xiongwei and Fu, Xiaojin and Zhong, Tao and Zeng, Jia and Song, Mingli and Chen, Gang},
  year = {2024}
}

@inproceedings{xiaoSurveyOptimizationModeling2025,
  title = {A survey of optimization modeling meets LLMs: progress and future directions},
  booktitle = {Proc. of IJCAI},
  author = {Xiao, Ziyang and Xie, Jingrong and Xu, Lilin and Guan, Shisi and Zhu, Jingyan and Han, Xiongwei and Fu, Xiaojin and Yu, WingYin and Wu, Han and Shi, Wei and Kang, Qingcan and Duan, Jiahui and Zhong, Tao and Yuan, Mingxuan and Zeng, Jia and Wang, Yuan and Chen, Gang and Zhang, Dongxiang},
  year = {2025},
  pages = {10742--10750}
}

@inproceedings{xieMURKAMultiRewardReinforcement2025,
  title = {MURKA: Multi-Reward Reinforcement Learning with Knowledge Alignment for Optimization Tasks},
  booktitle = {Proc. of NeurIPS},
  author = {Xie, Wantong and Hu, Yi-Xiang and Xu, Jieyang and Wu, Feng and Li, Xiangyang},
  year = {2025},
  pages = {34878--34905}
}

@article{yangAutomatedOptimizationModeling2025,
  title = {ORThought: Benchmarking and automating logistics optimization modeling via structured LLM reasoning},
  journal = {Artificial Intelligence for Transportation},
  author = {Yang, Beinuo and Zhou, Qishen and Li, Junyi and Su, Chenxing and Angeloudis, Panagiotis and Hu, Simon},
  year = {2026},
  volume = {6},
  pages = {100059}
}

@inproceedings{yangLargeLanguageModels2023,
  title = {Large Language Models as Optimizers},
  booktitle = {Proc. of ICLR},
  author = {Yang, Chengrun and Wang, Xuezhi and Lu, Yifeng and Liu, Hanxiao and Le, Quoc V. and Zhou, Denny and Chen, Xinyun},
  year = {2024}
}

@inproceedings{yangOptiBenchMeetsReSocratic2024,
  title = {OptiBench Meets ReSocratic: Measure and Improve LLMs for Optimization Modeling},
  booktitle = {Proc. of ICLR},
  author = {Yang, Zhicheng and Wang, Yiwei and Huang, Yinya and Guo, Zhijiang and Shi, Wei and Han, Xiongwei and Feng, Liang and Song, Linqi and Liang, Xiaodan and Tang, Jing},
  year = {2025}
}

@article{zhangOptimizationModelingSemantic2025,
  title = {SAC-Opt: Semantic Anchors for Iterative Correction in Optimization Modeling},
  author = {Zhang, Yansen and Kang, Qingcan and Chen, Yujie and Wang, Yufei and Han, Xiongwei and Zhong, Tao and Yuan, Mingxuan and Ma, Chen},
  journal = {arXiv:2510.05115},
  year = {2025}
}

@inproceedings{zhangSolvingGeneralNaturalLanguageDescription2024,
  title = {Solving General Natural-Language-Description Optimization Problems with Large Language Models},
  booktitle = {Proc. of NAACL Industry Track},
  author = {Zhang, Jihai and Wang, Wei and Guo, Siyan and Wang, Li and Lin, Fangquan and Yang, Cheng and Yin, Wotao},
  year = {2024},
  pages = {483--490}
}

@article{energy,
  author={Krishnamurthy, Dheepak and Uckun, Canan and Zhou, Zhi and Thimmapuram, Prakash R. and Botterud, Audun},
  journal={IEEE TPWRS}, 
  title={Energy Storage Arbitrage Under Day-Ahead and Real-Time Price Uncertainty}, 
  year={2018},
  volume={33},
  number={1},
  pages={84--93},
}

@article{economics,
  title = {Supply chain design and optimization: Challenges and opportunities},
  author = {Garcia, Daniel J. and You, Fengqi},
  year = {2015},
  journal = {Computers \& Chemical Engineering},
  volume = {81},
  pages = {153--170}
}

@article{math,
  author = {Delgado, Erwin J. and Cabezas, Xavier and Martin-Barreiro, Carlos and Leiva, Víctor and Rojas, Fernando},
  title = {An Equity-Based Optimization Model to Solve the Location Problem for Healthcare Centers Applied to Hospital Beds and COVID-19 Vaccination},
  journal = {Mathematics},
  year = {2022},
  volume = {10},
  number = {11},
  pages = {1825}
}

@article{others,
  title = {An overview of the optimization modelling applications},
  journal = {Journal of Hydrology},
  volume = {466--467},
  pages = {167--182},
  year = {2012},
  author = {Ajay Singh}
}

@inproceedings{reflection,
 author = {Shinn, Noah and Cassano, Federico and Gopinath, Ashwin and Narasimhan, Karthik and Yao, Shunyu},
 booktitle = {Proc. of NeurIPS},
 pages = {8634--8652},
 title = {Reflexion: language agents with verbal reinforcement learning},
 year = {2023}
}

@ARTICLE{MCTS,
  author={Browne, Cameron B. and Powley, Edward and Whitehouse, Daniel and Lucas, Simon M. and Cowling, Peter I. and Rohlfshagen, Philipp and Tavener, Stephen and Perez, Diego and Samothrakis, Spyridon and Colton, Simon},
  journal={IEEE TCIAIG}, 
  title={A Survey of Monte Carlo Tree Search Methods}, 
  year={2012},
  volume={4},
  number={1},
  pages={1--43},
}

@article{qwen3technicalreport,
  title={Qwen3 Technical Report}, 
  author={An Yang and Anfeng Li and Baosong Yang and Beichen Zhang and Binyuan Hui and Bo Zheng and Bowen Yu and Chang Gao and Chengen Huang and Chenxu Lv and Chujie Zheng and Dayiheng Liu and Fan Zhou and Fei Huang and Feng Hu and Hao Ge and Haoran Wei and Huan Lin and Jialong Tang and Jian Yang and Jianhong Tu and Jianwei Zhang and Jianxin Yang and Jiaxi Yang and Jing Zhou and Jingren Zhou and Junyang Lin and Kai Dang and Keqin Bao and Kexin Yang and Le Yu and Lianghao Deng and Mei Li and Mingfeng Xue and Mingze Li and Pei Zhang and Peng Wang and Qin Zhu and Rui Men and Ruize Gao and Shixuan Liu and Shuang Luo and Tianhao Li and Tianyi Tang and Wenbiao Yin and Xingzhang Ren and Xinyu Wang and Xinyu Zhang and Xuancheng Ren and Yang Fan and Yang Su and Yichang Zhang and Yinger Zhang and Yu Wan and Yuqiong Liu and Zekun Wang and Zeyu Cui and Zhenru Zhang and Zhipeng Zhou and Zihan Qiu},
  year={2025},
  journal={arXiv:2505.09388}, 
}

@inproceedings{SFT,
  title = {How Abilities in Large Language Models are Affected by Supervised Fine-tuning Data Composition},
  booktitle = {Proc. of ACL},
  author = {Dong, Guanting and Yuan, Hongyi and Lu, Keming and Li, Chengpeng and Xue, Mingfeng and Liu, Dayiheng and Wang, Wei and Yuan, Zheng and Zhou, Chang and Zhou, Jingren},
  year = {2024},
  pages = {177--198},
}

@inproceedings{sunRethinkingRewardModeling2024,
  title = {Rethinking Reward Modeling in Preference-based Large Language Model Alignment},
  booktitle = {Proc. of ICLR},
  author = {Sun, Hao and Shen, Yunyi and Ton, Jean-Francois},
  year = {2025}
}

@inproceedings{wangSemGuardRealTimeSemantic2025,
  title = {SemGuard: Real-Time Semantic Evaluator for Correcting LLM-Generated Code},
  booktitle = {Proc. of ASE},
  author = {Wang, Qinglin and Sun, Zhihong and Wang, Ruyun and Huang, Tao and Jin, Zhi and Li, Ge and Lyu, Chen},
  year = {2025},
  pages={1919--1930}
}

@article{guSemanticbasedOptimizationApproach2025,
  title = {A Semantic-based Optimization Approach for Repairing LLMs: Case Study on Code Generation},
  author = {Gu, Jian and Aleti, Aldeida and Chen, Chunyang and Zhang, Hongyu},
  year = {2025},
  journal = {arXiv:2503.12899}
}

@inproceedings{yang2024supercorrect,
  title = {SuperCorrect: Advancing Small LLM Reasoning with Thought Template Distillation and Self-Correction},
  booktitle = {Proc. of ICLR},
  author = {Yang, Ling and Yu, Zhaochen and Zhang, Tianjun and Xu, Minkai and Gonzalez, Joseph E and Cui, Bin and Yan, Shuicheng},
  year = {2025}
}

@article{islam2024enhancing,
  title={Enhancing Source Code Security with LLMs: Demystifying The Challenges and Generating Reliable Repairs},
  author={Islam, Nafis Tanveer and Khoury, Joseph and Seong, Andrew and Bou-Harb, Elias and Najafirad, Peyman},
  journal={arXiv:2409.00571},
  year={2024}
}

@inproceedings{xingHumanAlignedEvaluation2024,
  title = {Towards Human-aligned Evaluation for Linear Programming Word Problems},
  booktitle = {Proc. of LREC-COLING},
  author = {Xing, Linzi and Wang, Xinglu and Feng, Yuxi and Fan, Zhenan and Xiong, Jing and Guo, Zhijiang and Fu, Xiaojin and Ramamonjison, Rindra and Mostajabdaveh, Mahdi and Han, Xiongwei and Zhou, Zirui and Zhang, Yong},
  year = {2024},
  pages = {16550--16556}
}

@inproceedings{zhangOptiVerseBenchmark2026,
  title = {{O}pti{V}erse: A Comprehensive Benchmark towards Optimization Problem Solving},
  booktitle = {Proc. of ACL Findings},
  author = {Zhang, Xinyu and Zhang, Boxuan and Wan, Yuchen and Zhang, Lingling and Yao, Yixing and Wei, Bifan and Wu, Yaqiang and Liu, Jun},
  year = {2026},
  pages = {3059--3073}
}

\clearpage
\appendix

\section{MCTS Algorithm}
\label{sec:appendix_mcts_algorithm}

\begin{algorithm}[h]
\small
\caption{Semantic-Guided MCTS Process}
\label{alg:mcts}
\begin{algorithmic}[1]
\Require Problem Description $P$, Semantic Reward Model $R$, Policy Network $\pi_\theta$, Budget $T_{max}$, Expansion Count $H$, Exploration Constant $\omega$, Reflection Rounds $K$
\Ensure Best Solver Code $C^*$

\State Initialize tree $\mathcal{T}$ with root node $v^{(0)}$

\For{$iter \gets 1$ to $T_{max}$}
    \State $v^{(i)} \gets v^{(0)}$ 
    \While{$v^{(i)}$ is not a leaf node}
        \State $v^{(i+1)} \gets \mathop{\arg\max}_{u \in children(v^{(i)})} \left( Q(u) + \omega \cdot \sqrt{\frac{\ln N(v^{(i)})}{N(u) + \epsilon}} \right)$
        \State $v^{(i)} \gets v^{(i+1)}$
    \EndWhile
    \State $\tilde{v} \gets v^{(i)}$

    \If{$\tilde{v}$ is not a terminal state} 
        \State Sample $\{v^{(i+1)}_1, \dots, v^{(i+1)}_H\}$ via $v^{(i+1)}_j \sim \pi_\theta(\cdot | \tilde{v}, P)$
        \State Add $\{v^{(i+1)}_1, \dots, v^{(i+1)}_H\}$ to $\mathcal{T}$ as children of $\tilde{v}$
        \State $v^* \gets$ Select the node with the highest LLM prior
    \Else
        \State $v^* \gets \tilde{v}$
    \EndIf

    \If{$v^*$ is not a terminal state} 
        \State $C \gets \Call{Rollout}{v^*, \pi_\theta}$
    \Else
        \State $C \gets$ Extract solver code from $v^*$
    \EndIf
    \For{$k \gets 1$ to $K$} 
        \State Execute $C$ and collect output
        \If{no syntax error}
            \State \textbf{break}
        \Else
            \State $C \gets \pi_\theta(C, \text{error\_msg}, P)$ 
        \EndIf
    \EndFor
    \State $\mathbb{I} \gets \mathbf{1}[\text{no syntax error}]$
    \State Score $S \gets \mathbb{I} \cdot R(P, C)$

    \State $v \gets v^*$ 
    \While{$v \neq \text{NULL}$}
        \State $N(v) \gets N(v) + 1$
        \State $Q(v) \gets Q(v) + \frac{S - Q(v)}{N(v)}$
        \State $v \gets Parent(v)$
    \EndWhile
\EndFor

\State $v \gets v^{(0)}$
\For{$k \gets 1$ to $3$}
    \State $v \gets \mathop{\arg\max}_{u \in children(v)} Q(u)$
\EndFor
\State $C^* \gets \text{Solver code of } v$
\State \Return $C^*$
\end{algorithmic}
\end{algorithm}

\section{Training and Implementation Details}
\label{sec:appendix_implementation_details}

\subsection{Semantic Mutation Operators}
\label{sec:mutation_operators}

Table~\ref{tab:mutation_operators} details the mutation operators used to construct hard semantic negatives for reward-model training. These operators are designed to preserve executable solver code whenever possible while perturbing the modeling intent. They cover three common sources of semantic error in optimization modeling: variable declarations, objective definitions, and constraints.

Variable-level mutations alter the admissible solution space by relaxing variable types or removing bounds. Objective-level mutations change what the solver optimizes, either by flipping the optimization sense, changing coefficient signs, or dropping objective terms. Constraint-level mutations modify feasibility logic through comparator changes, right-hand-side shifts, variable swaps, index mismatches, or complete constraint removal. After the mutation process, candidates that fail execution or are judged infeasible are filtered out, so the remaining hard negatives correspond to executable but semantically incorrect solver code.

\begin{table}[h]
\centering
\caption{Semantic mutation operators for hard negative construction.}
\label{tab:mutation_operators}
\resizebox{\columnwidth}{!}{
\begin{tabular}{l|l|l}
\hline
\textbf{Dimension} & \textbf{Operator} & \textbf{Implementation Logic} \\ \hline
\multirow{2}{*}{Variable} & Type Relax & Relax variable types (e.g., BINARY $\to$ CONTINUOUS) \\
 & Bound Remove & Remove lower/upper bounds \\ \hline
\multirow{3}{*}{Objective} & Sense Flip & Swap direction (e.g., MIN $\to$ MAX) \\
 & Coeff Noise & Flip sign of variable coefficient in objective \\
 & Term Drop & Remove a term from objective expression \\ \hline
\multirow{5}{*}{Constraint} & Comparator Flip & Swap inequality signs (e.g., $\ge \to \le$) \\
 & RHS Shift & Modify RHS constant $b$ to $b \times (1 \pm \epsilon)$ \\
 & Var Swap & Swap positions of two variables of same type \\
 & Index Mismatch & Swap iteration sets in loop constructs \\
 & Constraint Drop & Remove an entire constraint declaration \\ \hline
\end{tabular}%
}
\end{table}

\subsection{Hyperparameter Settings}
\label{sec:appendix_hyperparams}

Table \ref{tab:hyperparams} details the hyperparameter configurations for the \model framework to facilitate reproducibility. The settings cover the generator, the MCTS reasoning process, and the adaptive gating mechanism. The exploration constant $\omega$ regulates the balance between visiting new nodes and exploiting high-scoring branches. The node expansion size $H$ defines the breadth of candidate generation at each reasoning step. Additionally, the gating threshold $\tau$ serves as the confidence cutoff for triggering the hierarchical search. This parameter is critical for balancing inference accuracy and computational cost by filtering straightforward instances.

\begin{table}[h]
    \centering
    \caption{Hyperparameter configurations for \model.}
    \label{tab:hyperparams}
    \resizebox{0.7\linewidth}{!}{
    \begin{tabular}{l|c}
    \toprule
    \textbf{Parameter} & \textbf{Value} \\
    \midrule
    \multicolumn{2}{l}{\textit{Generator (LLM) Settings}} \\
    Temperature & 0.9 \\
    Top-p & 0.9 \\
    Max Tokens & 2048 \\
    \midrule
    \multicolumn{2}{l}{\textit{MCTS Settings}} \\
    Exploration Constant ($\omega$) & 1.414 \\
    Simulations & [4, 16] \\
    Node Expansion Size ($H$) & 4 \\
    Reward Scale & [0, 1] \\
    \midrule
    \multicolumn{2}{l}{\textit{Adaptive Gating Settings}} \\
    Gating Threshold ($\tau$) & 0.6 \\
    \bottomrule
    \end{tabular}
    }
\end{table}

\subsection{Reward Model Training Details}
\label{sec:appendix_rm_training}

The semantic reward model is initialized from Qwen3-4B-Instruct-2507. We train it on 3,000 optimization-modeling problems from OR-Instruct~\cite{huangORLMCustomizableFramework2025}, split into 2,700 training and 300 validation problems, which yield 8,254 positive--negative preference pairs. We verify that the reward-model training problems do not overlap with the evaluation benchmarks. Table~\ref{tab:rm_training_details} summarizes the main training configuration.

\begin{table}[h]
    \centering
    \caption{Detailed training configuration for the semantic reward model.}
    \label{tab:rm_training_details}
    \resizebox{\linewidth}{!}{%
    \begin{tabular}{lcccccc}
        \toprule
        \textbf{Base Model} & \makecell{\textbf{Train / Val}\\\textbf{Problems}} & \textbf{Pairs} & \textbf{Epochs} & \textbf{Batch} & \textbf{LR} & \makecell{\textbf{Max}\\\textbf{Len.}} \\
        \midrule
        Qwen3-4B-Instruct-2507 & 2700 / 300 & 8254 & 3 & 32 & $2\mathrm{e}{-5}$ & 2048 \\
        \bottomrule
    \end{tabular}%
    }
\end{table}

\section{Baseline Details}
\label{sec:appendix_baselines}

This section supplements the baseline descriptions in the experimental setup.

\subsection{Standard Methods}
Standard Prompting is a single-pass generation baseline. It directly prompts the policy model to produce solver code from the problem description, without self-revision, external tool feedback, fine-tuning, or search. Self-Reflection follows the Reflexion framework~\cite{reflection}, which improves a language agent through verbal feedback stored in memory rather than parameter updates. In our setting, the model reflects on previous execution or solver feedback and regenerates solver code in later attempts.

\subsection{Workflow Methods}
OptiMUS~\cite{ahmaditeshniziOptiMUS03UsingLarge2025} is a modular LLM-based system for modeling and solving (MI)LP problems from natural-language descriptions. It develops math models, writes and debugs solver code, evaluates generated solutions, and uses these evaluations to improve correctness and efficiency. Chain-of-Experts~\cite{xiaoChainofExpertsWhenLLMs2023} is a multi-agent framework for complex OR problems. It assigns agents to specialized OR roles and uses a conductor to coordinate forward thought construction and backward reflection. OptiTree~\cite{liuOptiTreeHierarchicalThoughts2025} performs tree search over a hierarchical taxonomy of OR problem types. It identifies simpler subproblems and synthesizes global modeling thoughts from the selected hierarchy. SAC-Opt~\cite{zhangOptimizationModelingSemantic2025} uses semantic anchors extracted from the original problem and reconstructed from generated code. It aligns the two sets of anchors and selectively corrects mismatched objectives or constraints.

\subsection{Fine-tuned Methods}
ORLM~\cite{huangORLMCustomizableFramework2025} trains open-weight LLMs for optimization modeling using OR-Instruct, a semi-automated data synthesis framework. The resulting models are specialized for translating natural-language OR problems into math models and solver code. SIRL~\cite{chenSolverInformedRLGrounding2025} applies solver-informed reinforcement learning with verifiable rewards. External optimization solvers evaluate executable code and the associated LP representation, providing rewards based on syntax, feasibility, and solution quality.

\subsection{Search-based Methods}
Autoformulator~\cite{astorgaAutoformulationMathematicalOptimization2025} formulates optimization models through MCTS over a hierarchical space of modeling choices. It uses symbolic pruning to remove redundant branches and LLM-based evaluation to score partial math models. SolverLLM~\cite{liSolverLLMLeveragingTestTime2025} is a training-free test-time scaling framework. It generates math models and solver code through MCTS, using dynamic expansion, prompt backpropagation, and uncertainty backpropagation to guide the search process.

\section{Model-Level Semantic Validation}
\label{sec:appendix_semantic_validation}

Objective-value matching is the standard evaluation protocol on optimization-modeling benchmarks, but it credits a semantically incorrect formulation whenever that formulation happens to attain the reference optimum. This section reports a complementary validation of \model at the formulation level.

\paratitle{Evaluation Protocol.} Following prior work on human-aligned evaluation of LP word problems~\cite{xingHumanAlignedEvaluation2024}, we convert the predicted and the reference LP model into attributed bipartite graphs and compute the graph edit distance (GED) between them. Variable nodes encode bounds and objective coefficients, constraint nodes encode constraint bounds, and edges encode variable--constraint coefficients. We report three metrics. \textbf{GED Acc.} counts a prediction as correct only when the edit distance is zero, which yields a strict lower bound on formulation correctness. \textbf{Obj Acc.} is the objective-matching metric used in the main experiments, which is an upper bound because a semantically wrong model can return the correct objective by coincidence. For instances on which the two criteria disagree, we manually audit the predicted and the reference formulation, and report the human-verified correctness rate as \textbf{Oracle Acc.}

\paratitle{Dataset.} The benchmarks used in the main experiments either release only target objective values, contain too few formulation-level instances, or are too simple for the three metrics to separate meaningfully. We therefore extract 100 LP problems with explicit reference formulations from OptiVerse~\cite{zhangOptiVerseBenchmark2026}, denoted OptiVerse-LP-100, and compare \model against SIRL, the strongest fine-tuned baseline.

\begin{table}[h]
    \centering
    \caption{Formulation-level validation on OptiVerse-LP-100.}
    \label{tab:semantic_validation}
    \resizebox{0.85\linewidth}{!}{%
    \begin{tabular}{lccc}
        \toprule
        \textbf{Method} & \textbf{GED Acc.} & \textbf{Oracle Acc.} & \textbf{Obj Acc.} \\
        \midrule
        SIRL & 36\% & 36\% & 38\% \\
        \textbf{\model} & \textbf{44\%} & \textbf{47\%} & \textbf{49\%} \\
        \bottomrule
    \end{tabular}%
    }
\end{table}

\paratitle{Results.} As shown in Table~\ref{tab:semantic_validation}, \model improves over SIRL by 8 points in GED Acc., 11 points in Oracle Acc., and 11 points in Obj Acc. The gain therefore extends to formulation faithfulness rather than final-answer matching alone. The small gap between GED Acc. and Obj Acc. also indicates that objective matching is a useful proxy for semantic correctness, although not a complete one.

The two residual gaps have different causes. The Obj--Oracle gap corresponds to false positives under objective matching. In OptiVerse-53, for instance, the problem specifies a total investment of 200{,}000 yuan, whereas \model models it as an upper bound of at most 200{,}000 yuan; the optimum happens to exhaust the full budget, so the objective matches although the formulation does not. The GED--Oracle gap instead usually reflects imperfections in the reference model. In OptiVerse-678, the reference model imposes only non-negativity on the number of trucks, while \model additionally imposes integrality; GED marks this as a mismatch even though the stricter formulation is semantically valid.

\paratitle{Counterfactual Sensitivity.} We further test whether the reward model responds to specific problem statements rather than to surface features of the solver code. In ComplexLP-1, the original description requires food quantities to be integers, and the corresponding solver code encodes this integrality requirement, receiving a reward score of 0.72. We then modify only the problem description so that food quantities may be fractional, leaving the solver code unchanged. The reward score drops to 0.41. The reward model is thus sensitive to mismatches between individual problem requirements and the solver code, even though it emits a scalar score without sentence-level rationales.

\section{Parameter Sensitivity Analysis}

\subsection{Gating Threshold Sensitivity}
\label{sec:appendix_gating_sensitivity}

We tune the gating threshold $\tau$ on the validation set over $\{0.2,0.4,0.6,0.8,1.0\}$ and select $\tau=0.6$ as the default trade-off between accuracy and inference cost. Table~\ref{tab:gating_threshold_sensitivity} reports the corresponding test-set behavior on NL4Opt, ComplexLP, and ComplexOR. Lower thresholds trigger search only for low-confidence cases and reduce API calls, but they under-correct difficult instances. Higher thresholds approach the \textit{w/o Gating} setting and slightly improve accuracy on some datasets, but require substantially more calls. The default $\tau=0.6$ keeps accuracy close to the best setting while avoiding much of the extra inference cost.

\begin{table}[h]
    \centering
    \caption{Sensitivity to the gating threshold $\tau$. Each cell reports accuracy / API calls.}
    \label{tab:gating_threshold_sensitivity}
    \resizebox{\linewidth}{!}{%
    \begin{tabular}{lccccc}
        \toprule
        \textbf{Dataset} & $\tau=0.2$ & $\tau=0.4$ & $\tau=0.6$ & $\tau=0.8$ & $\tau=1.0$ \\
        \midrule
        NL4Opt & 72.9\% / 4.1 & 79.0\% / 8.4 & 87.9\% / 10.2 & 88.8\% / 19.6 & 88.8\% / 26.8 \\
        ComplexLP & 53.2\% / 4.6 & 73.8\% / 10.2 & 82.9\% / 13.6 & 83.8\% / 20.1 & 81.1\% / 27.4 \\
        ComplexOR & 44.4\% / 5.8 & 63.0\% / 13.1 & 66.7\% / 15.9 & 66.7\% / 22.3 & 66.7\% / 29.5 \\
        \bottomrule
    \end{tabular}%
    }
\end{table}

\subsection{Sampling Efficiency}
\label{sec:appendix_sampling_efficiency}

\begin{figure}[h]
    \centering
    \includegraphics[width=\linewidth]{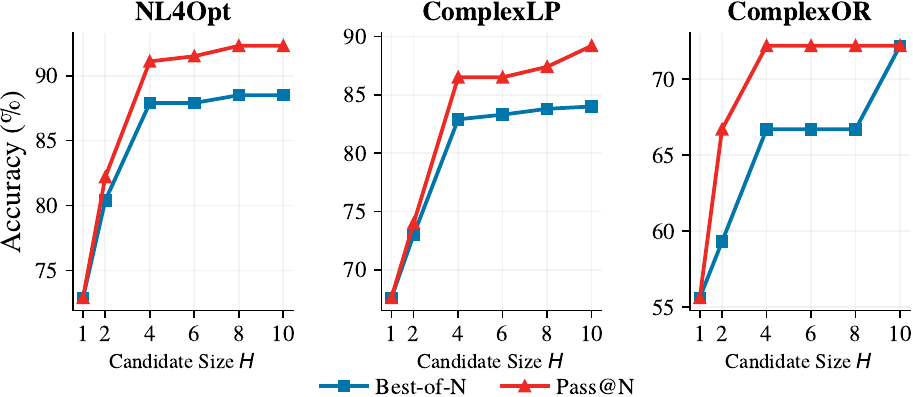}
    \caption{Impact of node expansion size $H$ on \model performance across three benchmarks.}
    \label{fig:sampling_efficiency}
\end{figure}

We further analyze the impact of the node expansion size $H$ on the performance of \model. As illustrated in Figure \ref{fig:sampling_efficiency}, we evaluate both Best-of-N and Pass@N metrics across NL4Opt, ComplexLP, and ComplexOR. The results indicate that while accuracy consistently improves as $H$ increases from 1 to 10, the performance gain saturates rapidly. Specifically, the accuracy typically reaches a plateau around $H=4$ across all datasets. This observation demonstrates the high sampling efficiency of our framework. It indicates that \model achieves near-optimal performance with a minimal computational budget, justifying our selection of $H=4$ as the default setting for the main experiments as reasonable.

\section{Result Variability Analysis}
\label{sec:appendix_hard_subset_variability}

\subsection{Run-to-Run Variation}
\label{sec:appendix_run_to_run_variation}

To assess the stability of the hard-subset comparison, Table~\ref{tab:hard_subset_variability} reports mean accuracy with run-to-run variation on the hard subsets of NL4Opt, ComplexLP, and ComplexOR. Each value is computed over repeated runs on the same instances; the reported $\pm$ values therefore describe variation induced by stochastic generation and MCTS exploration, rather than uncertainty over the benchmark population. \model improves over the strongest baseline on all three hard subsets, with the largest margin on ComplexOR. The relatively higher variation on NL4Opt reflects the fact that several instances lie near the decision boundary, so small changes in the sampled search trajectory can change the selected solution. Despite this variance, the mean accuracy of \model remains above the strongest baseline in every setting, confirming that the improvement is not attributable to a single favorable run.

\begin{table}[h]
    \centering
    \caption{Hard-subset accuracy with variation.}
    \label{tab:hard_subset_variability}
    \resizebox{0.85\linewidth}{!}{%
    \begin{tabular}{lcc}
        \toprule
        \textbf{Hard Subset} & \textbf{SIRL} & \textbf{\model} \\
        \midrule
        NL4Opt & 69.2\% $\pm$ 2.56\% & \textbf{71.8\%} $\pm$ 4.44\% \\
        ComplexLP & 54.5\% $\pm$ 5.25\% & \textbf{63.6\%} $\pm$ 3.03\% \\
        ComplexOR & 28.6\% $\pm$ 0.00\% & \textbf{57.1\%} $\pm$ 0.00\% \\
        \bottomrule
    \end{tabular}%
    }
\end{table}

\subsection{Paired Hard-Subset Analysis}
\label{sec:appendix_paired_hard_subset}

Run-to-run variation alone does not quantify uncertainty caused by the finite number of test instances. We therefore report absolute counts and paired instance-level tests in Table~\ref{tab:paired_hard_subset}. The confidence interval is a 90\% conditional exact interval for the paired net gain, computed from discordant pairs and scaled by the number of instances. The $p$-value is from the exact two-sided McNemar/binomial test on the same discordant outcomes.

\begin{table}[h]
    \centering
    \caption{Paired analysis on hard subsets. Counts are the number of solved instances over the subset size.}
    \label{tab:paired_hard_subset}
    \resizebox{\linewidth}{!}{%
    \begin{tabular}{lccccc}
        \toprule
        \textbf{Subset} & \textbf{SIRL} & \textbf{\model} & \makecell{\textbf{\model-only}\\\textbf{/ SIRL-only}} & \textbf{Gain} & \makecell{\textbf{90\% CI}\\\textbf{/$p$}} \\
        \midrule
        NL4Opt-hard & 27/39 & 28/39 & 1/0 & +2.6\% & $[-2.3\%,2.6\%]$ / 1.000 \\
        ComplexLP-hard & 18/33 & 21/33 & 4/1 & +9.1\% & $[-4.8\%,14.8\%]$ / 0.375 \\
        ComplexOR-hard & 2/7 & 4/7 & 2/0 & +28.5\% & $[-15.8\%,28.6\%]$ / 0.500 \\
        Combined hard & 47/79 & 53/79 & 7/1 & +7.6\% & $[0.6\%,10.0\%]$ / 0.070 \\
        \bottomrule
    \end{tabular}%
    }
\end{table}

The combined hard subsets contain 79 instances, on which \model wins seven paired cases and loses one, yielding a 7.6-point net gain and a positive 90\% interval. The ComplexOR-hard result is based on only seven instances (2/7 versus 4/7), so it should be interpreted as supportive rather than conclusive evidence. Taken together with the repeated-run results above, the paired counts provide a more complete picture: the advantage persists across the combined hard set, while the small ComplexOR subset warrants caution.

\section{Error Analysis}

\subsection{Syntactic Error Handling}
\label{sec:appendix_syntactic_errors}

This section specifies how syntactic and runtime errors are accounted for during search. When a rollout reaches a solver-code leaf, the code is executed. If the solver or the Python runtime reports an error, the error message is appended to the context and the generator regenerates the code for up to $K$ reflection rounds. If execution still fails after $K$ rounds, the rollout receives a score of $0$ and is not passed to the semantic reward model.Every node that initially triggers an error undergoes regeneration, and its final score enters backpropagation like any other node.

\begin{table}[h]
    \centering
    \caption{Syntactic error handling during search. The error rate is computed over all evaluated nodes; repair outcomes are computed over the nodes that initially triggered repair.}
    \label{tab:syntax_error_handling}
    \resizebox{\linewidth}{!}{%
    \begin{tabular}{lccc}
        \toprule
        \textbf{Dataset} & \makecell{\textbf{Syntax-error}\\\textbf{Rate}} & \makecell{\textbf{Correctly}\\\textbf{Repaired}} & \makecell{\textbf{Still Erroneous}\\\textbf{(Score 0)}} \\
        \midrule
        IndustryOR & 33.1\% & 64.1\% & 35.9\% \\
        ComplexLP & 34.2\% & 74.7\% & 25.3\% \\
        ComplexOR & 46.8\% & 81.8\% & 18.2\% \\
        \bottomrule
    \end{tabular}%
    }
\end{table}

Table~\ref{tab:syntax_error_handling} reports the resulting statistics. Roughly one third of the evaluated nodes on IndustryOR and ComplexLP trigger at least one execution error, and the rate rises to 46.8\% on ComplexOR, whose problems involve larger and more heavily indexed models. Reflection repairs the majority of these cases, and the remaining nodes are assigned a score of $0$. This separates code-level execution handling, which relies on solver feedback, from semantic correction, which is applied only to code that already runs.

\subsection{Semantic Error Taxonomy}
\label{sec:appendix_error_taxonomy}

To characterize the errors that remain after correction, we analyze the 40 instances that \model still answers incorrectly under BoN on IndustryOR, ComplexLP, and ComplexOR. We first categorize each error by the layer of the search hierarchy at which it originates, then further categorize the math-model errors by the component of the formulation that is wrong.

\begin{table}[h]
    \centering
    \caption{Distribution of the 40 remaining \model errors by hierarchy layer, and of the 28 math-model errors by formulation component.}
    \label{tab:error_taxonomy}
    \resizebox{0.9\linewidth}{!}{%
    \begin{tabular}{lcc}
        \toprule
        \textbf{Error Category} & \textbf{Count} & \textbf{Share} \\
        \midrule
        \multicolumn{3}{l}{\emph{By hierarchy layer (40 errors)}} \\
        Modeling Strategy & 9 & 22.5\% \\
        Math Model & 28 & 70.0\% \\
        Solver Code & 3 & 7.5\% \\
        \midrule
        \multicolumn{3}{l}{\emph{By formulation component (28 math-model errors)}} \\
        Parameter & 2 & 7.1\% \\
        Decision Variable & 7 & 25.0\% \\
        Objective Function & 4 & 14.3\% \\
        Constraint & 15 & 53.6\% \\
        \bottomrule
    \end{tabular}%
    }
\end{table}

As Table~\ref{tab:error_taxonomy} shows, most residual errors occur at the math-model layer, and constraints are the single largest source within that layer. This is consistent with the observation that constraints carry the implicit requirements of a problem description and are therefore the hardest component to recover from natural language. Only 7.5\% of the errors are introduced at the solver-code layer, indicating that implementation is rarely the bottleneck once a faithful math model is available. The 22.5\% of errors that originate at the strategy layer support including an explicit strategy layer in the search hierarchy, since an incorrect optimization framework or variable definition cannot be repaired by later layers. The taxonomy also suggests a practical ordering for future improvements: stronger extraction of implicit constraints should address the dominant error source, while better strategy classification can prevent errors from propagating into every downstream representation. In contrast, additional code-level repair is likely to have a smaller effect once the generated program is already executable.

\section{Qualitative Analysis}
\label{sec:case_study}

\begin{figure}[h]
    \centering
    \includegraphics[width=\linewidth]{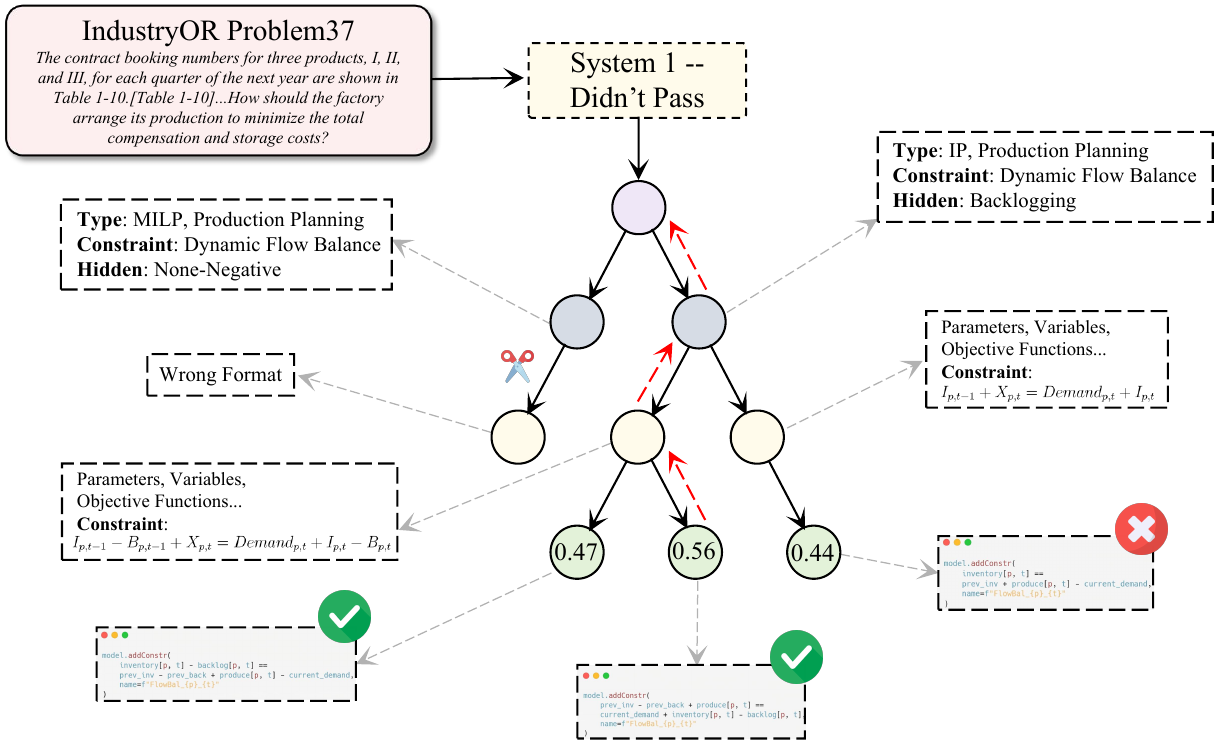}
    \caption{Case study on IndustryOR-37.}
    \label{fig:case_study}
\end{figure}

As illustrated in Figure \ref{fig:case_study}, we present a case study on ``IndustryOR-37'' to illustrate how \model resolves semantic ambiguities. The core challenge of this problem involves an implicit ``backlogging'' constraint. This allows for unmet demand to be carried over to subsequent periods, a logic that contradicts standard non-negative inventory assumptions.

\textbf{Failure of Direct Inference.} As shown in the figure, the direct inference (System 1) fails to capture this hidden requirement. It generates a standard inventory balance equation that assumes non-negativity, leading to an invalid model.

\textbf{Reasoning Process of \model.} In contrast, through its capacity of semantic discrimination, \model manages to figure out the correct answer. \model activates MCTS to explore the modeling space. At the leaf nodes, the semantic reward model distinguishes between plausible but incorrect math models. As shown in the figure, the model assigns the highest confidence score of 0.56 to the math model that explicitly incorporates the backlog variable. In contrast, it assigns a lower score of 0.44 to the candidate that omits this variable. This effectively suppresses the semantically misaligned solution. Both candidates are executable, so solver feedback alone cannot distinguish them; the decision depends on whether the formulation captures the problem's implicit backlog logic. This trajectory confirms that \model uses the reward model to actively reason through conflicting constraints and successfully rectifies semantic errors that standard prompting methods fail to identify.

\section{Prompts for Optimization}
\label{sec:prompts}

To ensure reproducibility, we provide the prompts of \model as follows:

\begin{figure*}[p]
\centering
\begin{promptbox}[title=Prompt for MCTS Layer 1: Modeling Strategy]
You are an expert in Operations Research. Your task is to determine the modeling strategy for the given problem.

\textbf{Problem Description:} \{problem\_description\}

\textbf{Your Task:}
1. Identify the specific optimization framework (e.g., LP, MILP, NLP).
2. Define the decision variables, specifying their types (Continuous, Integer, Binary) and semantic meaning.
3. List implicit constraints and key modeling assumptions.

\textbf{Output Format:}
Return a strictly valid JSON object with the following keys:
- \texttt{"problem\_type"}: The algorithmic framework (e.g., "Mixed Integer Linear Programming").
- \texttt{"decision\_variables"}: A list of objects, each with \texttt{"name"}, \texttt{"type"}, and \texttt{"description"}.
- \texttt{"assumptions"}: A list of strings clarifying ambiguities or implicit logic (e.g., "X must be non-negative").
- \texttt{"strategy\_summary"}: A concise high-level approach (no formulas).

\textbf{Guidelines:}
- \textbf{Variable Types:} Be rigorous. Use INTEGER/BINARY for discrete entities (people, counts) and CONTINUOUS for rates/amounts.
- \textbf{Framework:} Choose MILP if *any* discrete variables or logical constraints (if-then) are present.
- Output ONLY the JSON object.
\end{promptbox}
\vspace{-6pt}
\caption{Full prompt for the modeling-strategy layer of MCTS.}
\label{fig:prompt_strategy}
\end{figure*}

\begin{figure*}[p]
\centering
\begin{promptbox}[title=Prompt for MCTS Layer 2: Math Model]
You are an expert in mathematical optimization modeling. Translate the provided strategy into a clean LaTeX-style math model.

\textbf{Context:}
- Original problem: \{original\_problem\}
- Problem type: \{strategy.problem\_type\}
- Decision Variables: \{strategy.decision\_variables\}
- Assumptions: \{strategy.assumptions\}
- Strategy Summary: \{strategy.strategy\_summary\}

\textbf{Task Definition:}
1. Define all sets, parameters, variable declarations, objective, and constraint expressions based strictly on the provided strategy.
2. You must strictly adhere to the "Decision Variables" defined in the strategy snapshot. Do not change their types or meanings.
3. Preserve LaTeX notation, maintain consistent units, and ensure every implied constraint appears explicitly.

\textbf{Output Format:}
Provide the model as a single JSON object with exactly these five keys: \texttt{"sets"}, \texttt{"parameters"}, \texttt{"variables"}, \texttt{"objective"}, and \texttt{"constraints"}.

\textbf{Key Reminders:}
- Output only the JSON object---no explanatory text.
- Keep each field non-empty; add "None" explicitly if nothing applies.
- Ensure notation and units stay consistent throughout the math model.
\end{promptbox}
\vspace{-6pt}
\caption{Full prompt for the math model layer of MCTS.}
\label{fig:prompt_formulation}
\end{figure*}

\begin{figure*}[p]
\centering
\begin{promptbox}[title=Prompt for MCTS Layer 3: Solver Code]
You are an expert in optimization programming with Gurobi. Your task is to implement the provided math model as executable Python code.

\textbf{Context:}
- Original Problem: \{original\_problem\}
- Strategy Snapshot:
\ \ - Problem Type: \{strategy.problem\_type\}
\  \ - Decision Variables: \{strategy.decision\_variables\}
- Math Model: \{math\_model\}

\textbf{Task Definition:}
1. Import necessary libraries (\texttt{gurobipy}).
2. Define sample data for all parameters and sets using reasonable example values. The code must be directly executable.
3. Create the Gurobi model and define decision variables, strictly adhering to the types (Continuous, Binary, Integer) specified in the Strategy and Math Model.
4. Implement the objective function and all constraints exactly as formulated in the Math Model.
5. Perform the optimization (\texttt{model.optimize()}).
6. Include error handling for \texttt{GRB.INFEASIBLE} or \texttt{GRB.UNBOUNDED} statuses.
7. \textbf{CRITICAL:} If the solution is optimal, print the objective value using EXACTLY this format: \texttt{"Optimal objective value: " + str(model.objVal)}.

\textbf{Output Format:}
Provide ONLY the Python code. Do not include markdown formatting or explanatory prose.
\end{promptbox}
\vspace{-6pt}
\caption{Full prompt for the solver-code layer of MCTS.}
\label{fig:prompt_solver_code}
\end{figure*}

\clearpage

\end{document}